\documentclass{article}
\usepackage[letterpaper, margin=1in]{geometry}

\usepackage[utf8]{inputenc}
\usepackage[T1]{fontenc}

\usepackage{amsmath}
\usepackage{amssymb}
\usepackage{mathdots}
\usepackage{pgfplots}
\pgfplotsset{compat=1.18}

\usepackage{comment}

\usepackage[
  backend=biber,
  giveninits=true,
  maxbibnames=999,
  maxcitenames=999
]{biblatex}
\DeclareFieldFormat[misc]{title}{\mkbibquote{#1}}
\DeclareFieldFormat[misc]{citetitle}{\mkbibquote{#1}}
\renewbibmacro{in:}{}
\usepackage{hyperref}
\usepackage{cleveref} 

\crefname{equation}{}{}
\crefname{section}{\S}{\S\S}

\newcommand\Eta{\mathrm{H}}
\newcommand\Alpha{\mathrm{A}}
\newcommand\Beta{\mathrm{B}}

\newcommand\reals{\mathbb{R}}

\newcommand\cD{\mathcal{D}}
\newcommand\cU{\mathcal{U}}
\newcommand\cV{\mathcal{V}}
\newcommand\cM{\mathcal{M}}
\newcommand\cG{\mathcal{G}}
\newcommand\cGG{\bar{\cG}}
\newcommand\cA{\mathcal{A}}
\newcommand\cB{\mathcal{B}}
\newcommand\cW{\mathcal{W}}
\newcommand\cP{\mathcal{P}}
\newcommand\cMM{\bar{\cM}}
\newcommand\cT{\mathcal{T}}

\newcommand\from{\colon}

\DeclareMathOperator\rank{rank}

\DeclareMathOperator\Tr{tr}

\DeclareMathOperator\Retr{R}
\DeclareMathOperator\VT{VT}
\DeclareMathOperator\Skew{skew}
\DeclareMathOperator\Sym{sym}
\newcommand\St{\mathrm{St}}

\newcommand\LRq{\eta^{\mathrm{QKVO}}_*}
\newcommand\LRo{\eta^{\mathrm{other}}_*}

\newcommand\REPO{\url{https://github.com/nick-knight/low-rank-optimizers}}

\title{Riemannian Gradient Descent for Low-Rank Architectures}

\author{Nicholas Knight \\ NVIDIA \\ \texttt{nknight@nvidia.com}}
\date{22-May 2026}

\begin{document}
\maketitle

\begin{abstract}
We explore Riemannian optimization techniques for rank-factored matrix parameters,
targeting contemporary deep learning applications.
We examine ten points in the algorithm design space:
two geometries for rank-$r$ matrices,
three geometries for rank-$r$ partial isometries,
and block-matrix variants of these five,
where factors are shared across block-rows and block-columns.
We apply our methods to the multihead attention parameters in small language models.
After tuning learning rates,
our methods do not conclusively outperform an AdamW baseline.
Our implementations are available online\footnote{\REPO}.
\end{abstract}

\section{Introduction}

In modern machine learning applications,
almost every model parameter is a matrix or tensor entry (\emph{weight}).
Practitioners are thus incentivized to expose and exploit relationships between weights.
One such relationship is the rank of a matrix or tensor.

A popular way to constrain the rank of a weight matrix $W \in \reals^{m \times n}$ is to represent it in factored form, $W = A \cdot B^T$:
the rank of $W$ is bounded by the inner dimension $r$ of this matrix product.
Whereas $W$ has $mn$ parameters,
the pair $(A,B)$ has $mr+nr$,
potentially many fewer.
As a concrete example,
in multi-head attention~\cite{V+17},
the computations of each head $h$ involve two weight matrices,
$W_{QK,h}=W_{Q,h} \cdot W^T_{K,h}$
and
$W_{VO,h}=W_{V,h} \cdot W^T_{O,h}$,
both represented in factored form.
In contemporary instantiations,
the outer dimensions are often in the thousands, say 4096,
while the inner dimensions are often in the hundreds, say 128;
in this example, low-rank representations reduce the number of parameters $16$-fold.

In deep learning applications,
gradient-based training algorithms typically update $A$ and $B$ independently,
taking steps along their associated Euclidean gradients;
their product $W = AB^T$ plays no explicit role.
Here we take the opposite perspective: our goal is to learn a rank-constrained matrix $W$,
and the parameterization $AB^T$ is just an implementation detail.
Since the set $\reals^{m\times n}_r$ of rank-$r$ matrices is a smooth submanifold of $\reals^{m \times n}$,
it is possible to perform gradient descent on this submanifold;
this constrains rank automatically.
We consider several variants of gradient descent in this setting.

Our motivation for exploring manifold methods is heuristic.
Informally,
our broader assumption is that by better accounting for relationships between parameters,
an optimizer can converge in fewer iterations, or to a better solution.
For example,
note that $A\cdot B^T = AS \cdot S^{-1} B^T$ for any invertible $S$:
this AB-factorization is not unique.
Intuitively,
by focusing on $W$,
we can avoid meaningless search directions,
including those that lead us toward unbounded or uncharted regions.
Of course, we must always expect some cost-accuracy tradeoff,
so it's crucial to keep an eye on the per-iteration cost.
To this end,
all of our methods avoid forming or manipulating $m$-by-$n$ matrices like $W$,
and keep the per-iteration complexity $O(mr^2+nr^2)$.
Still, conventional approaches with per-iteration complexity $O(mr+nr)$ may remain the better choice,
even if they must run for more iterations.

Our main contribution is demonstrating application of Riemannian gradient descent to architectures with rank-factored matrix parameters,
like multihead attention.
Our algorithmic development started from the excellent textbooks of Absil, Mahony, and Sepulchre~\cite{AMS07} and Boumal~\cite{B23},
then followed the references therein.
Any apparent novelty in the present work is almost surely a rediscovery.

In \cref{sec:RGD} we present Riemannian gradient descent schematically.
In \cref{sec:algs} we instantiate this template for two geometries.
In \cref{sec:algs-ortho}
we address certain orthogonality constraints via the same template,
considering three geometries.
In \cref{sec:algs-grid}
we consider weight sharing,
adapting all five of the previous geometries.
In \cref{sec:expt} we conduct some numerical experiments and discuss their results.
In \cref{sec:future} we review future directions for this work.

\section{Riemannian Gradient Descent: Design Space}
\label{sec:RGD}

In our target applications, model training involves a form of (stochastic) gradient descent,
where (approximate) gradients are obtained via back-propagation
and passed to an optimizer,
which updates the model parameters.
Our goal in this work is to restrict our modifications to just the optimizer.
Algorithmically, this leads us to an embedded approach,
where each optimizer invocation begins with a gradient defined with respect to an ambient manifold.

We will consider a small family of methods,
which we describe together, schematically, in \cref{sec:RGD-schematic}.

\subsection{Schematic Iteration}
\label{sec:RGD-schematic}

Let $f$ be a real-valued objective,
defined on our model parameters,
to be minimized.
As above, let $\reals^{m\times n}_r$ denote the smooth submanifold of $\reals^{m \times n}$
containing the matrices of rank $r$.
We suppose both manifolds are endowed with Riemannian metrics:
the usual (Frobenius) inner product for the ambient manifold --- i.e., Euclidean geometry ---
and a currently unspecified geometry for the submanifold.
Let $W \in \reals^{m\times n}_r$ be a particular rank-constrained weight matrix in our model.
Restricting our attention to updating just $W$,
our schematic Riemannian gradient descent iteration is as follows.
\begin{enumerate}
\item Obtain $G_W$, the ambient (Euclidean) gradient of $f$ w.r.t.\ $W$.
\item Derive the associated Riemannian gradient, $G^R_W$. (This is a submanifold tangent vector.)
\item Augment our momentum, $M \gets \nu M + (1-\nu)G^R_W$, and determine our next step, $T = -\eta \cdot \mathrm{normalize}(M)$.
\item Move along $T$'s geodesic, from $W$ to $W'$, our next iterate. (This is the exponential map.)
\item Transport $M$ along the same geodesic, obtaining an element $M'$ of the tangent space at $W'$, used as the momentum in our next iteration. (This is parallel transport.)
\end{enumerate}

We have opted to include two features we call momentum and normalization,
which are not part of traditional gradient descent,
but are common in contemporary deep learning applications.

Conceptually, momentum biases the next step $T$ in the direction of previous steps;
we control this in Step~3, with parameter $\nu \in [0,1)$.
When momentum is disabled ($\nu = 0$), Step~5 can be omitted.

Normalization (Step~3) scales the next step candidate so its magnitude equals the learning rate $\eta > 0$.
Our primary interest is the norm induced by the Riemannian metric.
In practice we restrict to a trust region,
dividing by $\max(\| M\|, c)$ for some $c > 0$,
rather than by the raw norm $\|M\|$.
A different approach to normalization
is steepest descent in normed spaces;
we elaborate on this connection in \cref{sec:finsler}.

We have opted not to include weight decay,
another popular technique often combined with the previous ones.
We discuss weight decay,
along with other forms of momentum,
in \cref{sec:momentum}.

\section{Proposed Algorithms}
\label{sec:algs}

Let us now instantiate the Riemannian gradient descent template in \cref{sec:RGD}.
In \cref{sec:recover-GW} we address Step~1,
deriving the Euclidean gradient of $f$ w.r.t.\ $W$ from a factored representation.
The remaining steps are specific to our choice of submanifold geometry.
First we review manifold facts and establish notation in \cref{sec:prelim-manifold}.
Then
in \cref{sec:embedded,sec:quotient},
we instantiate our schematic for the fixed-rank matrix manifold $\reals^{m\times n}_r$, with \emph{embedded} and \emph{quotient} geometries.
In particular,
we only specify the Riemannian metric, and our approximations of the exponential map and parallel transport.
From a theoretical perspective,
all other details are fully specified by the Riemannian metric.
We defer practical implementation details to \cref{sec:impl}.

\subsection{Recovering the Euclidean Gradient}
\label{sec:recover-GW}

In our target applications,
$W$ is parameterized in factored form,
$W = AB^T$.
As our goal is to focus on the optimizer,
we will work within this parameterization;
relaxing this constraint is discussed in \cref{sec:parameterizations}.
As such, each invocation to our optimizer receives matrices $G_A$ and $G_B$,
the (Euclidean) gradients of $f$ w.r.t.\ $A$ and $B$.
Our task is to recover $G_W$, the gradient of $f$ w.r.t.\ $W$, from $G_A$ and $G_B$. This is Step~1, above.
Unfortunately, $G_W$ may not exist, and even if it does, it may not be uniquely defined by $G_A$ and $G_B$.

In this work, we assume that the objective $f$ depends only on $W=AB^T$,
not the factors $A$ and $B$ separately;
this is a sufficient condition for $G_W$ to exist.
(We discuss implications and relaxations of this assumption in \cref{sec:separable-loss}.)
Moreover,
under this assumption,
non-uniqueness turns out to be innocuous:
in all cases,
we obtain the same Riemannian gradient $G^R_W$ regardless of which version of $G_W$ we use.

All of our methods combine obtaining $G_W$ from $G_A$ and $G_B$ (Step~1) with deriving $G^R_W$ (Step~2),
thus avoiding forming $G_W$ explicitly.
Since $G^R_W$ is completely defined by the Riemannian metric,
we treat its computation as an implementation detail.

\subsection{Preliminaries}
\label{sec:prelim-manifold}

The textbook~\cite{B23} provides a self-contained reference for almost all of the following facts.

Let $\reals^{m\times n}$ denote the set of $m$-by-$n$ real-valued matrices.
Depending on context,
we interpret $\reals^{m\times n}$ as a smooth manifold or as a vector space,
each in the usual ways.
For each $W \in \reals^{m \times n}$, we identify $\cT_W \reals^{m\times n}$, the tangent space of (the manifold) $\reals^{m\times n}$ at $W$, with (the vector space) $\reals^{m\times n}$.
All manifolds we consider are derived from $\reals^{m\times n}$, with various $m,n$, via subsets, products, and quotients;
their topological and smooth structures,
and tangent space identifications,
are carried across these set-theoretic operations in the natural ways.
Hereafter, we understand all manifolds to be smooth, and all submanifolds to be embedded.

When $m,n$ are given, let $r^* = \min(m,n)$.
For each $r = 0,\ldots,r^*$,
let $\reals^{m\times n}_r$ be the submanifold of the rank-$r$ matrices,
abbreviating $\reals^{m\times n}_{r^*} = \reals^{m\times n}_*$.
Every $W \in \reals^{m \times n}_r$ can be represented,
nonuniquely,
as $W = AB^T$ for some $(A,B)$ in the product manifold $\reals^{m\times r}_* \times \reals^{n\times r}_*$.
Two AB-representations are equivalent, $(A,B) \sim (A',B')$,
iff $A' = AS$ and $B' = BS^{-T}$ for some $S \in \reals^{r \times r}_*$.
The quotient manifold $\reals^{m\times r}_* \times \reals^{n\times r}_* / \sim$
is diffeomorphic to $\reals^{m \times n}_r$,
and both the quotient map and the mapping $(A,B) \mapsto AB^T$ are smooth surjective submersions;
see, e.g.,~\cite[\S2]{AAM14}.
Hereafter, all submersions are smooth and surjective.

Let $\cM$ be any of our manifolds, with Riemannian metric $g$.
For any $W \in \cM$ and $\Xi \in \cT_W\cM$,
the exponential map takes a unit step from $W$ along the geodesic associated with $\Xi$.
(If $\cM$ is geodesically incomplete,
$\Xi$ is restricted to a star-shaped subset of $\cT_W\cM$.)
The exponential map is the solution of an initial value problem,
which we approximate crudely using retractions.
A retraction is a smooth map $\Retr \from \cD \to \cM$,
where $\cD$ is some neighborhood of the zero section of $\cT\cM$,
such that
the curve $c \from t \mapsto \Retr_W(t\Xi)$ satisfies $c(0) = W$ and $c'(0)=\Xi$.
Given any other $\Eta \in \cT_W\cM$,
parallel transport moves $\Eta$ along the aforementioned geodesic,
remaining parallel to the Levi-Civita connection.
This is again an initial value problem,
which we approximate using vector transports.
A vector transport associated with the retraction $\Retr$ is a smooth map $\VT \from \cD \times_\cM \cT\cM \to \cT\cM$ such that
$\VT_{W,\Xi}$ is a linear map from $\cT_W$ to $\cT_{\Retr_W(\Xi)}\cM$ and
$\VT_{W,0} = \mathrm{id}$.
We simply state our retractions and vector transports below,
deferring accuracy discussion to \cref{sec:retractions}.

Hereafter,
we abbreviate `Riemannian metric' simply as `metric'.
Similarly,
we understand `isometry' in the Riemannian sense,
with `linear isometry' meaning the maps between inner product spaces.

\subsection{Embedded Geometry}
\label{sec:embedded}

Our first metric on $\reals^{m\times n}_r$ is induced by endowing the ambient manifold $\reals^{m\times n}$ with the Frobenius inner product $\langle \cdot,\cdot\rangle$.
We call this the embedded geometry.
We learned of it from Boumal~\cite{B23},
who mainly credits Vandereycken~\cite{V13},
who in turn acknowledges numerous related investigations,
many appearing to be concurrent with his,
perhaps~\cite{SWC10} being the most immediately relevant.

Fix any $W \in \reals^{m\times n}_r$ and any $\Xi,\Eta \in \cT_W\reals^{m\times n}_r$.
We use the metric projection retraction $\Retr_W(\Xi) = U_r\Sigma_rV_r^T$,
where $(U_r,\Sigma_r,V_r)$ is a rank-$r$-truncated SVD of $W + \Xi$.
This matrix sum is well defined due to the inclusions established in \cref{sec:prelim-manifold}.
That $U_r\Sigma_rV_r^T$ exists as a rank-$r$ matrix, and is unique,
is equivalent to $\sigma_r > \sigma_{r+1}$,
which,
for theoretical purposes,
can be ensured by restricting $\|\Xi\|_2 < \sigma_r(W)/2$~\cite[Prop.~3.3]{AM12}.
(In practice, we retract first and ask questions later.)
We use the vector transport $\VT_{W,\Xi}(\Eta) = \Pi_{\Retr_W(\Xi)}(\Eta)$,
where $\Pi_W(Z) = AA^+Z + ZBB^+ - AA^+ZBB^+$,
for any AB-representation $(A,B)$ of $W$.
In words, we reinterpret $\Eta$ as an ambient tangent vector at the retracted point,
then project it orthogonally onto the submanifold tangent space.

We discuss our concrete implementation in~\cref{sec:impl-embedded}.
There is a decent amount of linear algebraic prestidigitation to avoid $m$-by-$n$ matrix operations.
Additionally,
we discuss several lurking numerical issues,
the most ominous one related to running off the end of a geodesic.

\subsection{Quotient Geometry}
\label{sec:quotient}

Recall from above that the mapping $(A,B) \mapsto AB^T$ is a submersion from
$\cMM=\reals^{m\times r}_* \times \reals^{n\times r}_*$
to $\cM = \reals^{m\times n}_r$.
Define the following metric ``upstairs'' on $\cMM$: for each $(A,B) \in \cMM$,
\[
\bar{g}_{(A,B)}\big((\Xi_A,\Xi_B),(\Eta_A,\Eta_B)\big)
=
\langle \Xi_A, \Eta_A\rangle_{B^TB} + \langle \Xi_B, \Eta_B \rangle_{A^TA}\text,
\]
where inner product weights are post-multiplied.
Note that, for all $S \in \reals^{r\times r}_*$,
\[
\bar{g}_{(A,B)}\big((\Xi_A,\Xi_B),(\Eta_A,\Eta_B)\big)
=
\bar{g}_{(AS,BS^{-T})}\big((\Xi_AS,\Xi_BS^{-T}),(\Eta_AS,\Eta_BS^{-T})\big)
\text,
\]
so this metric pushes forward through the submersion to a metric $g$ ``downstairs'' on $\cM$.
This gives a linear isometry between each tangent space downstairs,
and the horizontal subspace of each tangent space upstairs,
letting us emulate downstairs tangent-space operations via their (unique) horizontal lifts upstairs.
Hereafter we can focus on the upstairs operations.

Fix any $W \in \cM$ and any $(A,B) \in \cMM$ such that $W=AB^T$.
For any $\Xi,\Eta \in \cT_W\cM$, let $(\Xi_A,\Xi_B),(\Eta_A,\Eta_B)\in\cT_{(A,B)}\cMM$ be their horizontal lifts.
We use the additive retraction $\Retr_{(A,B)}((\Xi_A,\Xi_B)) = (A+\Xi_A,B+\Xi_B)$.
The result is rank-$r$ for sufficiently small $\Xi$.
For example,
by Weyl's inequality, it suffices that $\|\Xi_A\|_2 < \sigma_r(A)$ and $\|\Xi_B\|_2 < \sigma_r(B)$.
For lagniappe, we'll give a slightly stronger sufficient condition, that both $\|A^+\Xi_A\|_2$ and $\|B^+\Xi_B\|_2$ are less than one.
Informally, a large perturbation need not threaten column rank if it's orthogonal to the column space.
Let's check this.

Consider the A-factor first.
If $\|A^+ \Xi_A\|_2 < 1$,
then every eigenvalue $\lambda_i$ of $A^+\Xi_A$ has $|\lambda_i|<1$,
thus $\det(I + A^+\Xi) = \prod_i (1 + \lambda_i) \ne 0$,
so $I + A^+\Xi_A$ is invertible.
Toward a contradiction,
suppose $\rank(A + \Xi_A) < r$,
i.e., $(A+\Xi_A)x = 0$ for some $0 \ne x \in \reals^r$.
Premultiplying by $A^+$,
we must then conclude that $(I + A^+\Xi_A)x = 0$,
but this contradicts the invertibility just established,
thus $A + \Xi_A$ must be full rank.
Having established sufficiency for the A-factor,
note that $\|A^+\Xi_A\|_2 \le \|A^+\|_2 \|\Xi_A\|_2 = \|\Xi_A\|_2 / \sigma_r(A)$,
since $\|\cdot\|_2$ is submultiplicative and $\sigma_1(A^+) = 1/\sigma_r(A)$ by the definition of the pseudoinverse,
so $\|A^+ \Xi_A\|_2 < 1$ implies the aforementioned Weyl-based condition.
The same reasoning applies to the B-factor.

We use the vector transport $\VT_{(A,B),(\Xi_A,\Xi_B)}((\Eta_A,\Eta_B)) = (\Eta_A-A'\Lambda, \Eta_B+B'\Lambda^T)$,
where $(A',B')$ denotes the retracted point and $\Lambda = ({A'}^+\Eta_A - \Eta_B^T {B'}^{+T})/2$.
In words, we reinterpret the horizontal lift of $\Eta$ as a tangent vector (upstairs) at the retracted point,
then project it horizontally there.

Our choice of the metric $\bar{g}$ was rather heuristic:
it was the simplest factor-space metric we could find that had the desired invariance under~$\sim$.
We found it in~\cite{MAS12}.
It also turned out to be amenable to a rather simple implementation, as shown in \cref{sec:impl-quotient}.
We mention a reasonable alternative in \cref{sec:parameterizations}.

\section{Extension: Orthogonality Constraints}
\label{sec:algs-ortho}

We extend our approach to further constrain the rank-$r$ matrix $W=AB^T$ to be a partial isometry.
This means that all $r$ of $W$'s nonzero singular values equal one;
equivalently, $A$ and $B$ are column-orthonormal.
The set of such matrices is a submanifold of $\reals^{m\times n}_r$,
and the Riemannian approach applies.
As with the rank constraint,
we must ensure this new constraint is met at parameter initialization,
but need make no changes to the model architecture.

Let's expand our notation from \cref{sec:prelim-manifold} with two new submanifolds of $\reals^{m\times n}$.
Let $\St_{m,n}$ be the (Stiefel) submanifold of matrices with orthonormal columns ($m \ge n$) or rows ($m \le n$).
%
For all $r \le r^*$, let $\cP^{m,n}_r$ be the submanifold of rank-$r$ partial isometries.
(The term `partial isometry' is only ever used in the present context.)
Every $W \in \cP^{m,n}_r$ can be represented, nonuniquely, as $W = UV^T$ for some $(U,
V)$ in the product manifold $\St_{m,r} \times \St_{n,r}$.
Two UV-representations are equivalent,
$(U,V)\sim (U',V')$,
iff $U'=UQ$ and $V'=VQ$ for some $Q \in \St_{r,r}$.
The quotient manifold $\St_{m,r} \times \St_{n,r} / \sim$ is diffeomorphic to $\cP^{m,n}_r$,
and both the quotient map and the mapping $(U,V) \mapsto UV^T$ are submersions.

In \cref{sec:embedded-ortho,sec:quotient-ortho,sec:canonical},
we instantiate our schematic for the partial isometry manifold $\cP^{m,n}_r$ with \emph{embedded}, \emph{quotient}, and \emph{canonical} geometries.

\subsection{Embedded Geometry for Partial Isometries}
\label{sec:embedded-ortho}

For the embedded geometry,
we proceed as in \cref{sec:embedded},
further restricting the ambient Frobenius inner product to $\cP^{m,n}_r$.

Fix any $W\in \cP^{m,n}_r$, and any $\Xi,\Eta \in \cT_W \cP^{m,n}_r$.
As before, we use the metric projection retraction $\Retr_W(\Xi) = U_rV_r^T$,
where $(U_r,\Sigma_r,V_r)$ is a rank-$r$ truncated SVD of $W+\Xi$.
Specializing the fixed-rank case, we conclude this is well defined if $\|\Xi\|_2 < 1/2$.

Our vector transport is $\Pi_{\Retr_W(\Xi)}(\Eta)$,
where
\[ \Pi_W(Z) = (I-UU^T)ZVV^T + UU^TZ(I-VV^T) + U \Skew(U^T Z V) V^T\text, \]
with $\Skew(M) = (M - M^T)/2$, for any UV-representation $(U,V)$ of $W$.

Our implementation is detailed in \cref{sec:impl-embedded-ortho}.

\subsection{Quotient Geometry for Partial Isometries}
\label{sec:quotient-ortho}

For the quotient geometry,
we proceed similarly to \cref{sec:quotient},
with $(U,V)\mapsto UV^T$ our submersion from
$\cMM = \St_{m,r} \times \St_{n,r}$ to $\cM=\cP^{m,n}_r$.
Our upstairs metric $\bar{g}$ is just the restriction of the one in \cref{sec:quotient};
however, the inner product weights disappear due to orthonormality,
so $\bar{g}$ coincides with (a restriction of) the product Frobenius metric.
That is, for each $(U,V) \in \cMM$,
\[ \bar{g}_{(U,V)}\big((\Xi_U,\Xi_V),(\Eta_U,\Eta_V)\big)=
\langle \Xi_U, \Eta_U\rangle + \langle \Xi_V, \Eta_V \rangle\text;\]
moreover, for all $Q \in \St_{r,r}$,
\[
\bar{g}_{(U,V)}\big((\Xi_U,\Xi_V),(\Eta_U,\Eta_V)\big)
=
\bar{g}_{(UQ,VQ)}\big((\Xi_UQ,\Xi_VQ),(\Eta_UQ,\Eta_VQ)\big)
\text,
\]
leading to a Riemannian submersion as before.

Fix any $W \in \cM$ and any $(U,V) \in \cMM$ such that $W = UV^T$.
For any $\Xi,\Eta \in \cT_W\cM$, let $(\Xi_U,\Xi_V),(\Eta_U,\Eta_V)\in\cT_{(U,V)}\cMM$ be their horizontal lifts.
We consider two retractions $\Retr_{(U,V)}((\Xi_U,\Xi_V))$,
which are standard Stiefel retractions performed factorwise.
For the QR retraction, we return the Q-factors from the thin QRs of $U+\Xi_U$ and $V+\Xi_V$, both constrained so that the R-factors have nonnegative diagonals.
For the polar retraction, we return the column-orthonormal factors from the polar decompositions of $U+\Xi_U$ and $V+\Xi_V$.
(Other retraction options are discussed in \cref{sec:retractions}.)
Both these retractions are well defined if their inputs are full rank,
which is guaranteed by orthogonality and tangency.
That is,
$(U + \Xi_U)^T(U + \Xi_U) = I + \Xi_U^T \Xi_U$,
since $U^TU = I$ and $U^T \Xi_U = -\Xi_U^T U$,
and this matrix is symmetric positive definite,
verifying that $U + \Xi_U$ is full rank;
the same reasoning applies to $V + \Xi_V$.

Our vector transport first reinterprets $(\Eta_U,\Eta_V)$ as an ambient tangent vector at the retracted point $(U',V')$,
then projects it onto $\cT_{(U',V')} \cMM$,
then projects it further onto the horizontal subspace.
Altogether,
\[
\VT_{(U,V),(\Xi_U,\Xi_V)}((\Eta_U,\Eta_V)) =
(T_U - U' \Omega, T_V - V' \Omega)
\]
where
\begin{align*}
T_U &= \Eta_U - U' \Sym ({U'}^T \Eta_U)
\\
T_V &= \Eta_V - V' \Sym ({V'}^T \Eta_V)
\\
 \Omega &= ({U'}^T T_U + {V'}^T T_V)/2
\text,
\end{align*}
where $\Sym(M) = (M+M^T)/2$.
Our implementation is detailed in \cref{sec:impl-quotient-ortho}.

\subsection{Canonical Geometry for Partial Isometries}
\label{sec:canonical}
We repeat \cref{sec:quotient-ortho},
but instead of specializing $g$ from \cref{sec:quotient},
we endow each of the Stiefel factors with the `canonical' metric~\cite{EDS98},
giving $\cMM$ the product metric.
That is, for each $(U,V) \in \cMM$,
\[ \bar{g}_{(U,V)}\big((\Xi_U,\Xi_V),(\Eta_U,\Eta_V)\big)=
\langle \Xi_U, \Eta_U\rangle_{I-UU^T/2} + \langle \Xi_V, \Eta_V \rangle_{I-VV^T/2}\text,\]
where inner product weights are pre-multiplied.
This satisfies the same orthogonal invariance as the previous metric,
making the submersion Riemannian once again.

%
We use the same retraction and vector transport as before.
Indeed, even though the metric changes from \cref{sec:quotient-ortho},
the horizontal spaces and projections turn out to be the same,
enabling substantial logic reuse.
Our implementation is detailed in \cref{sec:impl-canonical}.

\section{Extension: Weight Sharing}
\label{sec:algs-grid}

Now
we address a model-architectural design that is presently quite popular:
sharing factor matrices across multiple rank-factored parameters.
For example, in group-query attention~\cite{GQA},
the $H$ heads are divided into $G$ groups,
where each group $g$ shares $W_{K,g}$ and $W_{V,g}$,
while each head $h$, regardless of group, still has its own $W_{Q,h}$ and $W_{O,h}$.
(The special case $G=1$, proposed earlier~\cite{MQA}, is usually called multi-query attention, but we won't make this distinction.)
More examples and discussion follows in \cref{sec:tensor}.

We consider a two-dimensional generalization of the preceding example,
where we have a collection $(W_{ij})_{ij}$ of rank-$r$ matrix parameters, each factoring as $W_{ij} = A_i B^T_j$.
We sometimes arrange these as submatrices of a larger block matrix $\cW = \cA \cB^T$,
where $\cA$ and $\cB$ stack the factors $A_i$ and $B_j$ vertically.
That is, each $A_i$ is shared across a block-row of $\cW$, and each $B_j$ down a block-column.
The set of such collections $(W_{ij})_{ij}$ is a submanifold of a product manifold, whose factors are copies of $\reals^{m\times n}_r$ for various conformable $m,n$;
the Riemannian approach again applies.
This further extends to the case treated in \cref{sec:algs-ortho}, where each $W_{ij}$ is a partial isometry.
We call these submanifolds \emph{grid manifolds}, in reference to the Cartesian weight-sharing pattern.

Conceptually, grid algorithms map to the schematic in \cref{sec:RGD},
reinterpreting objects like $W$ as collections $(W_{ij})_{ij}$.
The discussion in \cref{sec:recover-GW} also applies to the grid case,
substituting $\cW=\cA\cB^T$ for $W=AB^T$,
and assembling $G_\cW$, $G_\cA$, and $G_\cB$ from $(G_{W_{ij}})_{ij}$, $(G_{A_i})_i$, and $(G_{B_j})_j$ analogously.
Now, the key assumption is that $f$ is a function of each $W_{ij}$, not its factors $A_i$ and $B_j$ independently.
Additionally, all five geometries treated in \cref{sec:algs,sec:algs-ortho}
in the `singleton' case, naturally carry over to the `grid' case, via product metrics.
Generalizing the algorithms for those geometries, however, is less straightforward.

In the fixed-rank case (\cref{sec:algs}),
it turns out we can simply stack the factors as $\cA$ and $\cB$,
and stack their gradients similarly,
then invoke the singleton algorithms on these larger matrices.
We justify this in \cref{sec:grid-embedded,sec:grid-quotient}.
The main wrinkle is that we now need to ensure that the retractions preserve rank blockwise.

In the partial isometry case,
this stacking trick does not seem to work,
and our algorithms,
detailed in \cref{sec:grid-embedded-ortho,sec:grid-quotient-ortho,sec:grid-canonical},
are more complicated.

Before proceeding to the algorithms, let's again extend our notation from \cref{sec:prelim-manifold}.
Our grid manifolds are submanifolds of $\prod_{ij}\reals^{m_i \times n_j}$,
where $m_i,n_j \ge r$, and the indices $i$ and $j$ always run from $1$ through $I$ and $J$, resp.

In the fixed-rank case, the grid manifold is
\[
\big\{ (A_iB_j^T)_{ij} : A_i \in \reals^{m_i \times r}_*, B_j\in\reals^{n_j \times r}_*\big\}
\text.
\]
Analogous to the singleton case,
every $(W_{ij})_{ij}$ can be represented by some $((A_i)_i,(B_j)_j)$ in the product manifold
$\prod_i \reals^{m_i \times r}_* \times \prod_j \reals^{n_j \times r}_*$.
Nonuniqueness of these grid AB-representations is characterized by the equivalence relation,
$((A_i)_i,(B_j)_j) \sim ((A_iS)_i,(B_jS^{-T})_j)$ for all $S \in \reals^{r\times r}_*$,
leading to the grid mapping $((A_i)_i,(B_j)_j) \to (A_iB_j^T)_{ij}$ being a submersion.

In the partial isometry case, the grid manifold is
\[
\big\{ (U_iV_j^T)_{ij} : U_i \in \St_{m_i,r}, V_j\in\St_{n_j, r}\big\}
\text.
\]
Again, analogous to the singleton case,
every $(W_{ij})_{ij}$ can be represented by some $((U_i)_i,(V_j)_j)$ in the product manifold
$\prod_i \St_{m_i,r} \times \prod_j \St_{n_j,r}$.
Nonuniqueness of these grid UV-representations is characterized by the equivalence relation,
$((U_i)_i,(V_j)_j) \sim ((U_iQ)_i,(V_jQ)_j)$ for all $Q \in \St_{r,r}$,
leading to the grid mapping $((U_i)_i,(V_j)_j) \to (U_iV_j^T)_{ij}$ being a submersion.

\subsection{Fixed Rank, Embedded Geometry, Grid Case}
\label{sec:grid-embedded}

Continuing notation,
let $\cG$ denote our fixed-rank grid manifold,
a submanifold of $\prod_{i,j} \reals^{m_i \times n_j}$.
Endow $\prod_{i,j} \reals^{m_i \times n_j}$ with the product Frobenius metric,
which $\cG$ naturally inherits.

Let $m=\sum_i m_i$ and $n=\sum_j n_j$ and
let $\cG'$ denote the submanifold $\big\{\cW \in \reals^{m \times n}_r : \cW_{ij} \in \reals^{m_i \times n_j}_r \big\}$ of $\reals^{m \times n}_r$, itself a submanifold of $\reals^{m \times n}$.
Endow $\reals^{m \times n}$ with the Frobenius metric,
which $\reals^{m \times n}_r$ and $\cG'$ inherit.
The stacking transformation
$F:(W_{ij})_{ij} \mapsto \cW =\big[ W_{ij} \big]_{ij}$
is an isometry from $\prod_{i,j} \reals^{m_i \times n_j}$ to $\reals^{m \times n}$,
as well as from $\cG$ to $\cG'$.
Crucially, on top of being isometrically embedded, $\cG'$ is an open (i.e., codimension zero) submanifold of $\reals^{m \times n}_r$.

This all means that the manifold operations in the singleton case (\cref{sec:embedded})
applied to the stacked matrix $\cW = \cA \cB^T$,
are locally equivalent to ones on $\cG'$,
and thus on $(W_{ij})_{ij} \in \cG$.
In particular, the retraction and vector transport remain well defined
in a neighborhood $\cD$ of the zero section of $\cT\cG$.
However,
the upper bound on $\|\Xi\|_2$ from the singleton case does not ensure blockwise rank preservation.
In the grid case,
it suffices to ensure that $\|\Xi\|_2 < s/2$,
where $s=\min_{i,j} \sigma_r(W_{ij})$.
(Note that this recovers the singleton condition~\cite[Prop.~3.3]{AM12} when $I=J=1$.)
Let's derive this.

We want to show that every retracted block remains rank-$r$.
That is,
forming $R$ from any $r$-truncated SVD of $\cW + \Xi$,
we want to ensure that
$\sigma_r (R_{ij}) > 0$ for all $i,j$.
Now, for any $i,j$, we have that
\[
\sigma_r(R_{ij})
\ge
\sigma_r(W_{ij}) - \|R_{ij} - W_{ij}\|_2
\ge
\sigma_r(W_{ij}) - \|R - \cW\|_2
\ge s - \|R - \cW\|_2
\text,
\]
where the first inequality uses Weyl's inequality (see, e.g.,~\cite[Cor.~8.6.2]{GVL}),
and the second follows from Cauchy's interlacing theorem (see, e.g.,~\cite[Thm.~1]{T72}).
What remains is to enforce $\|R - \cW\|_2 < s$.
Now,
\[
\|(\cW+\Xi)-R\|_2 = \sigma_{r+1}(\cW + \Xi) \le \|\Xi\|_2
\text,
\]
due to the Eckart-Young-Mirsky theorem (see, e.g.,~\cite[Thm.~2.4.8]{GVL}),
in which sense $R$ is optimal,
and Weyl's inequality,
noting that $\sigma_{r+1}(\cW)=0$.
So, by the triangle inequality,
\[
\|R - \cW\|_2 \le \|(\cW + \Xi)-R\|_2 + \|\Xi\|_2 \le 2 \|\Xi\|_2
\text.
\]
Thus it suffices to ensure that $2 \|\Xi\|_2 < s$.

For the retraction to be well defined,
we also need to ensure that $R$ is unique.
Letting $S = \sigma_r(\cW)$,
Weyl tells us that $\sigma_r(\cW + \Xi) \ge S - \|\Xi\|_2$,
and that $\sigma_{r+1}(\cW+\Xi) \le \|\Xi\|_2$,
and Cauchy tells us that $s \le S$.
Thus $\|\Xi\|_2 < s/2$ implies that $\sigma_r(\cW+\Xi) > \sigma_{r+1}(\cW + \Xi)$.
This gap implies that, for any $r$-truncated SVD $(U_r,\Sigma_r,V_r)$,
the product $R=U_r\Sigma_rV^T_r$ is the same.

In \cref{sec:impl-grid-embedded},
we derive a sharper upper bound on $\|(\cW+\Xi) - R\|_2$
that yields the sufficient condition $\|\Xi\|_2 < \frac{sS}{s+S}$,
which coincides with the one above in the singleton case,
but is stronger in the grid case.

Implementation-wise,
the only real difference from the singleton case
is checking $\|\Xi\|_2$ or verifying ranks.

\subsection{Fixed Rank, Quotient Geometry, Grid Case}
\label{sec:grid-quotient}

Recall from above that the mapping $((A_i)_i,(B_j)_j) \mapsto (A_iB_j^T)_{ij}$ is a submersion from
$\cGG = \prod_i \reals^{m_i \times r}_* \times \prod_j \reals^{n_j \times r}_*$
to our grid fixed-rank manifold $\cG$.
Define the following metric on $\cGG$:
for each $((A_i)_i,(B_j)_j)\in\cGG$,
\[
\Big(\big((\Xi_{A_i})_i,(\Xi_{B_j})_j\big),\big((\Eta_{A_i})_i,(\Eta_{B_j})_j\big)\Big)\mapsto
\sum_{ij}
\langle \Xi_{A_i}, \Eta_{A_i}\rangle_{B_j^T B_j} + \langle \Xi_{B_j}, \Eta_{B_j} \rangle_{A_i^TA_i}
\text,
\]
where inner products are weighted by post-multiplication.
This metric is invariant under $\sim$,
so descends through the quotient to a metric downstairs on $\cG$,
thus the submersion $((A_i)_i,(B_j)_j) \mapsto (A_iB_j^T)_{ij}$ is Riemannian.

Letting $m=\sum_i m_i$ and $n=\sum_j n_j$,
reconstruct, following the singleton case (\cref{sec:quotient}), the Riemannian submersion
$(\cA,\cB)\mapsto \cA\cB^T$ from
$\cMM = \reals^{m\times r}_* \times \reals^{n\times r}_*$
to $\cM = \reals^{m\times n}_r$.
The same map is a Riemannian submersion from the submanifold
$\cGG' = \big\{(\cA,\cB) \in \cMM : \rank(\cA_i)=\rank(\cB_j) = r \big\}$ of $\cMM$
to the submanifold $\cG' = \big\{\cW \in \cM : \rank(\cW_{ij}) = r \big\}$ of $\cM$,
where the subscripts on $\cA_i$, $\cB_j$, and $\cW_{ij}$ extract blocks appropriately.
As in \cref{sec:grid-embedded}, these blockwise rank constraints make these both open submanifolds.

The stacking transformation $F \from ((A_i)_i, (B_j)_j) \mapsto (\cA,\cB)$
is a diffeomorphism from $\cGG$ to $\cGG'$,
and from $\cG$ to $\cG'$.
By construction of the metrics on $\cGG$ and $\cMM$,
$F$ is an isometry $\cGG$ to $\cGG'$;
moreover,
noting that $F$ is invariant under the equivalence relations on its domain and image,
that isometry descends to one between $\cG$ and $\cG'$.

This all means that the manifold operations from the singleton case (\cref{sec:quotient}),
applied to the stacked matrices $(\cA,\cB) \in \cGG'$,
are locally equivalent to ones on $\cA\cB^T \in \cG'$, and thus on $(A_iB_j^T)_{ij}\in\cG$.
As in the case of \cref{sec:grid-embedded},
the retraction and vector transport remain well defined
in a neighborhood $\cD$ of the zero section of $\cT\cG$.
Applying the argument from the singleton case blockwise,
it suffices to ensure that $\|A_i^+\Xi_{A_i}\|_2 < 1$ and $\|B_j^+\Xi_{B_j}\|_2 < 1$ for all $i,j$.

We remark on the metric we defined earlier on $\cGG$, which may have seemed a strange choice at the time.
For example, although we defined it on a product manifold,
it is not the product metric of (independent) metrics on the factors.
As we saw,
it turned out to be our familiar quotient metric on $\cMM$, as inherited by $\cGG'$ and pulled back through $F$,
a pragmatic choice to make our diagram commute.
Perhaps more satisfying, it happens to equal the product quotient metric on $\prod_{ij} \reals^{m_i \times r}_* \times \reals^{n_j \times r}_*$,
at the points that share AB-factors in the desired way, i.e., $(A_{ij},B_{ij})_{ij}$ such that $A_{ij} = A_{ij'}$ and $B_{ij}=B_{i'j}$ for all $i,i',j,j'$.

Again, implementation-wise, there are no major changes from the singleton case,
aside from needing more care to ensure the retraction preserves rank blockwise.

\subsection{Partial Isometry, Embedded Geometry, Grid Case}
\label{sec:grid-embedded-ortho}

Incorporating our orthogonality constraint,
our grid manifold is now $\cG = \big\{ (U_iV_j^T)_{ij} : U_i \in \St_{m_i,r}, V_j\in\St_{n_j,r}\big\}$.
As in the fixed-rank grid case (\cref{sec:grid-embedded}),
$\cG$ naturally inherits the product Frobenius metric from $\prod_{i,j} \reals^{m_i \times n_j}$.
Unlike that case,
the singleton algorithm (\cref{sec:embedded-ortho}) does not extend so straightforwardly.
We describe two approaches that we abandoned,
then the approach we ultimately took.

First, let's attempt to extend the stacking approach from \cref{sec:grid-embedded}.
Immediately there is a problem.
The stacked $\cW = \cU\cV^T$,
while rank-$r$,
is not a partial isometry when $IJ>1$:
in particular, $\cW$'s singular values all equal $\sqrt{IJ}$.
However, what we have is just a homothety of $\cP^{m,n}_r$:
in principle we can address this simply by scaling the metric, norm, geodesics, etc.
Following this path shortly leads us to a major obstruction.
Recall previously we found an isometry from the grid fixed-rank manifold to an open, isometrically embedded submanifold of some  (singleton) fixed-rank manifold.
Here too we find an isometry from the grid partial isometry manifold $\cG$ to an isometrically embedded submanifold of some scaled (singleton) partial isometry manifold.
However, this submanifold is not open, due to the blockwise Stiefel constraints $U_i^TU_i = V_j^TV_j = I$.
Even tiny steps could violate these constraints.
At this point we abandoned stacking, and started considering finer-grained approaches.

Our second attempt tried to address the grid geometry directly.
Obtaining the Riemannian gradient was somewhat tedious but routine,
and we were able to form a candidate step $(T_{ij})_{ij}$ in $\cT_{(W_{ij})_{ij}}\cG$
(see \cref{sec:impl-grid-embedded-ortho}).
Our metric projection retraction, however, now means minimizing
\[ \sum_{ij} \| W_{ij} + T_{ij} - U_iV_j^T\|^2_F \]
such that
$U_i \in \St_{m_i,r}$ and $V_j \in \St_{n_j,r}$ for all $i,j$.
When $IJ=1$, this has an SVD-based solution that we exploited in \cref{sec:embedded-ortho},
but we didn't see any obvious, comparably simple solution when $IJ>1$.
We decided to defer further investigation on this subproblem to future work,
and seek a less principled retraction that's easier to compute.

Briefly, we were working in parallel on the other two geometries for the grid partial isometry manifold $\cG$,
discussed next in \cref{sec:grid-quotient-ortho,sec:grid-canonical},
and things were easier there.
So what we ended up doing was jury-rigging our retraction and vector transport from those geometries to finish our embedded algorithm.
In particular,
we lift $(T_{ij})_{ij}$ to the product manifold $\cGG=\prod_i \St_{m_i,r} \times \prod_j \St_{n_j,r}$,
using the horizontal lift for the quotient geometry,
then perform factorwise polar retraction and vector transport via tangent projection and horizontal (re)projection for the quotient geometry.
Let us comment on this surprising turn of events.

First, why switch geometry when lifting?
In principle,
for any smooth splitting of the tangent bundle upstairs and any metric downstairs,
we can define a `compatible' metric upstairs.
Perhaps unsurprisingly,
we found all candidate upstairs metrics to be computationally unwieldy:
in particular, they didn't split nicely factorwise,
obfuscating an efficient retraction.
However, we did observe some structure in the inter-factor coupling,
which could probably be exploited.
Although we weren't convinced this was a dead end,
it didn't seem easier than working downstairs.

Second, is switching geometries like this even valid?
Indeed,
despite working upstairs in the quotient geometry,
the result downstairs is still a valid retraction and vector transport for the embedded geometry:
the requisite smoothness and linearity are preserved up through the lift,
horizontally via retraction/transport, then back down through the submersion.
We believe this is a special case of a more general property of submersions,
not necessarily Riemannian, not necessarily induced by Lie group actions,
and we hope to clarify this in future work.

Third, we mention that our preference for the polar retraction is due to its equivariance under right multiplication by $Q \in \St_{r,r}$,
meaning that the induced retraction and vector transport downstairs remain independent of the choice of representative upstairs.
The QR retraction, on the other hand, is generally not equivariant in this way,
and we were rather concerned about the practical ramifications,
even though the induced operation downstairs is technically a retraction.
This also warrants further study.

\subsection{Partial Isometry, Quotient Geometry, Grid Case}
\label{sec:grid-quotient-ortho}

The stacking approach encounters the same underlying difficulty we faced in the embedded case (\cref{sec:grid-embedded-ortho}):
the grid partial isometry manifold $\cG$ is diffeomorphic to a submanifold of a scaled (singleton) partial isometry manifold,
but this submanifold is not open, due to the blockwise Stiefel constraints $U_i^TU_i = V_j^TV_j = I$,
so there is no reason to expect operations on the singleton manifold to emulate, even locally, the desired ones on the grid manifold.
This problem is independent of geometry. Nor can we lift our way around it.
However, the techniques we developed in the singleton case do generalize here.

Recall from above that the mapping $((U_i)_i,(V_j)_j)\mapsto (U_iV_j^T)_{ij}$ is a submersion from $\cGG=\prod_i \St_{m_i,r} \times \prod_j \St_{n_j,r}$ to our grid partial isometry manifold $\cG$.
The metric from \cref{sec:grid-embedded} induces one on our present $\cGG$ via inclusion,
\[
\Big(\big((\Xi_{U_i})_i,(\Xi_{V_j})_j\big),\big((\Eta_{U_i})_i,(\Eta_{V_j})_j\big)\Big)\mapsto
J \sum_i \langle \Xi_{U_i}, \Eta_{U_i}\rangle
+
I\sum_j \langle \Xi_{V_j}, \Eta_{V_j} \rangle
\text.
\]
As in the singleton case, the inner-product weights cancel,
but unlike that case,
it does not end up perfectly coinciding with the product Euclidean metric,
due to the factors $J,I$ scaling the sums.
Conceptually, this weights each factor $U_i,V_j$ according to how many blocks $W_{ij}$ it influences.
Similarly to the grid fixed-rank quotient case,
another motivation for this choice is that it
coincides with the product quotient metric on $\prod_{ij} \St_{m_i,r} \times \St_{n_j,r}$ at points that share UV-factors in the desired way.

Fix any $(W_{ij})_{ij} \in \cG$ and any $((U_i)_i,(V_j)_j) \in \cGG$ such that $W_{ij} = U_iV_j^T$ for all $i,j$.
For any $\Xi,\Eta \in \cT_{(W_{ij})_{ij}}\cG$, let $((\Xi_{U_i})_i,(\Xi_{V_j})_j),((\Eta_{U_i})_i,(\Eta_{V_j})_j)\in\cT_{((U_i)_i,(V_j)_j)}\cGG$ be their horizontal lifts.
Our retractions are the same factorwise Stiefel ones, QR and polar, from the singleton case (\cref{sec:quotient-ortho}),
except now we have $I$ left factors and $J$ right factors,
but they remain well defined by the same reasoning.
Our vector transport is a straightforward generalization of the singleton case.
Let $((U'_i)_i,(V'_j)_j)$ denote the retracted point.
First, for all $i,j$, we compute the tangent projections,
\begin{align*}
T_{U_i} &= \Eta_{U_i} - U'_i \Sym ({U'_i}^T \Eta_{U_i})
\\
T_{V_j} &= \Eta_{V_j} - V'_j \Sym ({V'_j}^T \Eta_{V_j})
\end{align*}
then remove the vertical components,
\[
\VT_{((U_i)_i,(V_j)_j),((\Xi_{U_i})_i,(\Xi_{V_j})_j)}(((\Eta_{U_i})_i,(\Eta_{V_j})_j)) =
((T_{U_i} - U'_i \Omega)_i, (T_{V_j} - V'_j \Omega)_j)
\text,
\]
where $\Omega = \big(J\sum_i {U'_i}^T T_{U_i} + I \sum_j {V'_j}^T T_{V_j}\big)/(2IJ)$.

Our implementation is detailed in \cref{sec:impl-grid-quotient-ortho}.

\subsection{Partial Isometry, Canonical Geometry, Grid Case}
\label{sec:grid-canonical}

For the canonical geometry,
we use a similar scaling as in the preceding case;
our metric is
\[
\Big(\big((\Xi_{U_i})_i,(\Xi_{V_j})_j\big),\big((\Eta_{U_i})_i,(\Eta_{V_j})_j\big)\Big)\mapsto
J \sum_i \langle \Xi_{U_i}, \Eta_{U_i}\rangle_{I-U_iU_i^T/2}
+
I\sum_j \langle \Xi_{V_j}, \Eta_{V_j} \rangle_{I-V_jV_j^T/2}
\text,
\]
where inner product weights are pre-multiplied.
Like before,
this coincides with the product canonical metric on $\prod_{ij} \St_{m_i,r} \times \St_{n_j,r}$ under weight sharing.

As in the singleton case,
the horizontal spaces and projections turn out to be the same as in the quotient geometry,
and we use the same retractions and vector transport.

\section{Experiments}
\label{sec:expt}

Via smaller-scale language modeling experiments,
we conclude that our algorithms are functional,
and have potential to be competitive with contemporary defaults.
We focus on accuracy, as measured by a loss function,
and do not evaluate algorithmic costs (runtime, energy, etc.).
Our ongoing work seeks to provide more useful guidance for practitioners,
with competitive benchmarking at larger model scales.

\subsection{Setup}

Our language model architecture is a decoder-only variant of the `base' model in~\cite{V+17}:
it has 6 layers, 8 attention heads per layer,
and embedding dimension of 512,
so each attention head has dimension 64,
and each feedforward network has internal dimension 2048.
We modify the architecture in a few ways:
the input embedding and output unembedding are untied;
there are no positional encodings (relative or absolute);
normalization is root-mean-squared,
with no learnable parameters,
applied to the inputs of each self-attention and feedforward block;
and, the feedforward networks use squared-ReLU activations.
We consider two architectural variants:
the ``MHA'' model follows the original multihead attention approach (no weight sharing),
while the ``GQA'' model uses group-query attention (sharing of the K and V weights),
with query group size 8 (i.e., maximal sharing).
We apply our five singleton (non-grid) approaches to the MHA model,
and our five grid approaches to the GQA model.

Our objective,
a standard one in this application domain,
is to minimize the cross-entropy between the model's predicted next-token distribution and that of the sampled data.
We normalize the loss per-batch by the number of token predictions.
Our corpus is FineWeb~\cite{P24}, a publicly available set of documents distilled from years of Web crawls.
We use a curated subset called  `sample-10BT',
which we randomly shuffle at a document granularity and split into training and validation subsets.
We use the `GPT2' tokenizer from tiktoken.
Our dataloader processes each document in the (shuffled) order,
tokenizing it with an end-of-sequence token prepended,
extracting successive subsequences of 512 tokens,
treated thereafter as distinct samples.
Samples cross document boundaries to reach length 512.
Once 8 samples have been obtained,
we feed that batch to the model,
and discard any remainder of the document being processed.
Our loss calculation ignores the model's prediction for the token past the end of each sample.
Training and validation steps use the same sequence length, batch size, and loss calculation.

In our models, all learnable parameters are associated with a matrix.
Aside from the QKVO weights, we have
an input embedding,
an output unembedding,
and two fully connected layers for each feedforward (sub)network.
We initialize each matrix parameter's entries with i.i.d.\ samples from $U[-a, a]$,
where $a$ is the inverse square root of the dimension of the domain of the linear operator associated with the matrix.
We then orthogonalize the initial QKVO parameters,
even though this is only required for the partial isometry approaches.
We fix three seeds for our random number generator,
defining three different initializations for the MHA model, and for the GQA model.
Momentum buffers are always zero-initialized.

Despite our efforts to control stochasticity,
we did notice slight run-to-run variability in a few instances,
which we suspect was due to dynamic reassociation of floating-point arithmetic in certain math libraries.
We continue to investigate the root causes.

Our baseline is AdamW~\cite{LH19}, using PyTorch's default settings ($\beta_1=0.9$, $\beta_2 = 0.999$, $\epsilon=10^{-8}$, $\lambda=0.01$).
We use this optimizer for all model parameters,
selectively swapping it out with our low-rank optimizers just for the QKVO parameters.
We use two maximal learning rates,
$\LRq$, for the QKVO parameters, and
$\LRo$, for the non-QKVO parameters.
We warm up both learning rates linearly from zero over the first 10 training steps, and hold them constant thereafter.

We use bfloat16 for model parameters and forward/backward pass arithmetic.
We use float32 for optimizer states and optimizer arithmetic.
Our implementations are written in Python, using the PyTorch framework.
We use \texttt{torch.compile} to optimize the forward/backward pass and the optimizer step.
We run on a NVIDIA 8x H100 node, running CUDA 12.4, using NVIDIA's PyTorch 25.08 container.

\subsection{Procedure}

All experiments used a fixed batch size of 8.

We first optimized our AdamW baseline, by doing a grid search over learning rates $\LRq,\LRo = 2^{-k}$ for $k=14,12,10,8,6$.
We trained for 1000 steps,
validating every 10 steps.
For each run,
we averaged the validation loss over the last 5 validation steps,
then we averaged this over the three random seeds.
Our key findings were as follows:
\begin{itemize}
\item For both MHA and GQA models,
the optimal surfaces appeared strictly convex on this grid;
the two minima happened to coincide at $(\LRq,\LRo) = (2^{-12}, 2^{-10})$.
\item For all runs,
validation loss was ${\sim}11$ nats after the first training step,
and the best last-5 means were
6.391 nats ($\sigma = 0.010$) for MHA, and
6.388 nats ($\sigma = 0.005$) for GQA.
\item
Optimality appeared much more sensitive to $\LRo$ than to $\LRq$.
\end{itemize}

Keeping $\LRo=2^{-10}$,
we then swapped out AdamW on the QKVO parameters for each of our low-rank optimizers,
then swept $\LRq = 2^{-k}$, $k = 12,10,8,6,4,2,0$, toggling metric normalization.
For these experiments we clamped vanishing norms to $c = 2^{-23}$,
kept momentum disabled, and
only used polar retractions (where applicable).
The rest of the setup was the same as before.
Our key findings were as follows:
\begin{itemize}

\item
Metric normalization improved loss by about 0.01 nats in each MHA case,
and by about 0.05 nats in each GQA case.

\item
For each method, the
optimal surface appeared strictly convex on the $\LRq$ grid.
The optimal $\LRq$ was constant within MHA and GQA cases: $2^{-8}$ and $2^{-4}$, resp.

\item
In the metric normalization case,
for MHA, loss varied between 6.385--6.393 across methods at optimal $\LRq$,
and for GQA it varied between 6.378--6.383.
At each of these 10 points,
the standard deviations across the three seeds varied between 0.001-0.013 nats.
Within MHA and GQA cases,
the difference between geometries is within seed variance.
The improvements over the baseline are stronger in the GQA case,
but still borderline explainable by seed variance.

\end{itemize}

Next we performed a similar experiment targeting momentum,
comparing $\nu = 1/8$ and $\nu = 7/8$ with keeping momentum disabled ($\nu = 0$).
We kept metric normalization enabled (clamping to $2^{-23}$),
and for each of the 10 optimizers we tried three values of $\LRq$:
its optimum found in the previous experiment, times $1/2,1,2$.
Our key findings were as follows:
\begin{itemize}
\item
In the MHA case, all results were within the seed-noise floor of the configuration established in the previous experiment.
In the GQA case, there was a slight improvement from increasing to $\LRq=2^{-3}$,
but it was independent of changing $\nu$ from $0$ to $1/8$.

\item
For GQA fixed-embedded case, with the suboptimal $\LRq=2^{-5}$,
we saw a 0.015 nat benefit from $\nu = 7/8$,
which was above the noise floor.

\item
In all cases, heavy momentum hurts by 0.10--0.14 nats at the larger learning rate.

\end{itemize}

Finally, we performed a larger scaling study.
All experiments continued to use AdamW on the non-QKVO parameters with $\LRo = 2^{-10}$.
For the QKVO parameters, AdamW used $\LRq=2^{-12}$ (both MHA and GQA),
while the low-rank optimizers used $\LRq=2^{-8}$ (MHA) and $2^{-3}$ (GQA).
The low-rank optimizers used metric normalization (clamping to $2^{-23}$),
no momentum,
and polar retractions where applicable.
We ran for 10,000 steps, validating every 10.
We average training and validation loss over the three seeds.

\pgfplotsset{
  curve adamw/.style              ={color=black,           thick, solid},
  curve fixed embedded/.style     ={color=blue,            thick, solid},
  curve fixed quotient/.style     ={color=orange!90!black, thick, solid},
  curve partial embedded/.style   ={color=green!55!black,  thick, dashed},
  curve partial quotient/.style   ={color=red!85!black,    thick, dashed},
  curve partial canonical/.style  ={color=violet,          thick, dashed},
  longrun axis/.style={
    width=0.85\linewidth, height=0.32\linewidth,
    grid=major, grid style={dotted, gray!50},
    xlabel={training step},
    legend pos=outer north east,
    legend cell align=left,
    no markers,
    smooth,
  },
  longrun rel axis/.style={
    longrun axis,
    ymin=-0.4, ymax=0.4,
    ylabel={loss $-$ AdamW},
  },
}

\newcommand{\addlongruncurve}[4]{%
  \addplot[#1] table[x=step, y=#2, col sep=comma] {#4};%
  \addlegendentry{#3}%
}
\newcommand{\addrelcurve}[4]{%
  \addplot[#1] table[x=step, y expr=\thisrow{#2}-\thisrow{adamw}, col sep=comma] {#4};%
  \addlegendentry{#3}%
}

\newcommand{\longrunmainpanel}[2]{%
  \begin{tikzpicture}
    \begin{axis}[longrun axis, ylabel={#2}]
      \addlongruncurve{curve adamw}{adamw}{AdamW}{#1}
      \addlongruncurve{curve fixed embedded}{fixed_embedded}{fixed embedded}{#1}
      \addlongruncurve{curve fixed quotient}{fixed_quotient}{fixed quotient}{#1}
      \addlongruncurve{curve partial embedded}{partial_embedded}{partial embedded}{#1}
      \addlongruncurve{curve partial quotient}{partial_quotient}{partial quotient}{#1}
      \addlongruncurve{curve partial canonical}{partial_canonical}{partial canonical}{#1}
    \end{axis}
  \end{tikzpicture}%
}

\newcommand{\longrunrelpanel}[1]{%
  \begin{tikzpicture}
    \begin{axis}[longrun rel axis]
      \addrelcurve{curve adamw}{adamw}{AdamW}{#1}
      \addrelcurve{curve fixed embedded}{fixed_embedded}{fixed embedded}{#1}
      \addrelcurve{curve fixed quotient}{fixed_quotient}{fixed quotient}{#1}
      \addrelcurve{curve partial embedded}{partial_embedded}{partial embedded}{#1}
      \addrelcurve{curve partial quotient}{partial_quotient}{partial quotient}{#1}
      \addrelcurve{curve partial canonical}{partial_canonical}{partial canonical}{#1}
    \end{axis}
  \end{tikzpicture}%
}

\newcommand{\longrunfigurebody}[2]{%
  \longrunmainpanel{#1}{#2}\par\vspace{0.3em}%
  \longrunrelpanel{#1}%
}

\begin{figure}[t]
\centering
\longrunfigurebody{data/mha_train.csv}{training loss}
\caption{%
MHA training loss over 10\,000 steps (sampled every 20 steps),
averaged over three seeds.
Five `singleton' low-rank algorithms vs.\ AdamW.
\emph{Top:} full y-range.
\emph{Bottom:} relative to AdamW.
}%
\label{fig:MHA-train}
\end{figure}

\begin{figure}[t]
\centering
\longrunfigurebody{data/mha_valid.csv}{validation loss}
\caption{%
MHA validation loss associated with \cref{fig:MHA-train},
averaged over three seeds.
\emph{Top:} full y-range.
\emph{Bottom:} relative to AdamW.
}%
\label{fig:MHA-valid}
\end{figure}

\begin{figure}[t]
\centering
\longrunfigurebody{data/gqa_train.csv}{training loss}
\caption{%
GQA training loss over 10\,000 steps (sampled every 20 steps),
averaged over three seeds.
Five `grid' low-rank algorithms vs.\ AdamW.
\emph{Top:} full y-range.
\emph{Bottom:} relative to AdamW.
}%
\label{fig:GQA-train}
\end{figure}

\begin{figure}[t]
\centering
\longrunfigurebody{data/gqa_valid.csv}{validation loss}
\caption{%
GQA validation loss associated with \cref{fig:GQA-train},
averaged over three seeds.
\emph{Top:} full y-range.
\emph{Bottom:} relative to AdamW.
}%
\label{fig:GQA-valid}
\end{figure}

Training and validation losses for the MHA model are in \cref{fig:MHA-train,fig:MHA-valid},
and for the GQA model in \cref{fig:GQA-train,fig:GQA-valid}.
(For visualization purposes, we only plot training loss every 20 steps.)
Our key findings were as follows:
\begin{itemize}
\item For all runs, the training and validation loss curves appear qualitatively similar.

\item Two of the three MHA AdamW runs had sharp loss spikes around step 6660, reaching ${\sim}50\times$ loss, although they ultimately recovered.
Our fixed-quotient and partial-quotient GQA methods spiked up to ${\sim}1.5\times$ in the same window, and also recovered.
We examined the training data in the window around step 6660 and observed a relatively high concentration of structured documents: lists, Wikipedia tables, a Grateful Dead setlist, a recipe, etc.
We also reran the AdamW runs with weight decay disabled, and these loss spikes persisted, now also appearing in the GQA runs, and at several other steps.

\item One of the three GQA AdamW runs had a ${\sim}5\times$ loss spike around step 5100, but also recovered.

\item For MHA, we observe AdamW converging slightly faster than our methods, aside from the step-6660 spike.
Our embedded geometries seem to perform slightly better than the others.

\item For GQA, our methods follow AdamW much more closely, in some cases slightly outperforming it up until the step-6660 spike.

\end{itemize}

\subsection{Discussion}

When replacing AdamW with our low-rank optimizers for the QKVO parameters,
and tuning learning rate,
we observed roughly comparable training dynamics.
This supports the hypothesis that our algorithms are mathematically sound and our implementations are correct.
This also supports the hypothesis that our algorithms are competitive with AdamW in terms of data efficiency.
However,
the algorithmic complexities of our methods are substantially larger than AdamW's,
so from a practical perspective, our experiments do not support replacing AdamW.

These conclusions are overshadowed by several significant limitations of our experimental design.
Perhaps the biggest limitation is that our model and token horizon are quite small by modern standards.
It is unclear how training behavior may change at larger scales.
Another major limitation is our omission of learning rate annealing,
which is important for convergence, both in theory and in practice.
(Our ad hoc 10-step warmup, and similarly ad hoc batch-size choice,
also warrant revisiting.)
A third limitation is that we did not tune AdamW's parameters from their defaults,
or compare with other optimizers.

\section{Related and Future Work}
\label{sec:future}

A large community contributed to the theory we've invoked in the present work.
The textbooks~\cite{AMS07,B23} survey the vast and mature field of Riemannian optimization.
The book chapter~\cite{UV20} focuses specifically on applications to rank-structured matrices and tensors.

There are many different directions for future work.
We highlight a few directions that we find interesting or promising.

\subsection{Normed Descent}
\label{sec:finsler}

In 1944, Curry suggested performing steepest descent under arbitrary norms,
not necessarily induced by inner products~\cite{C44}.
This technique has been applied fruitfully throughout the modern deep learning era (see, e.g.,~\cite{BN24}).
For example, AdamW,
the reference optimizer we failed to beat experimentally,
is closely related to steepest descent under the $\ell_\infty$ norm.
And Muon~\cite{Muon},
a newer AdamW rival,
can be viewed as steepest descent, for matrix parameters, under the spectral (Schatten-$\infty$) norm.

Normed descent meets the present work in the context of Finsler manifolds,
whose geometry is induced by norms on the tangent spaces,
rather than by inner products,
like Riemannian manifolds.
Concretely,
the Finsler variant of our approach picks our search direction $\Xi \in \cT_W\reals^{m\times n}_r$ to minimize $df_W[\Xi]$ subject to $\|\Xi\|\le 1$,
for some norm $\|\cdot \|$;
then we perform retraction and vector transport.
(The exponential map and parallel transport have Finsler analogues, too.)
We briefly investigated this technique for Schatten-$p$ norms ($1 \le p \le \infty$).
The case $p=2$ is Riemannian, our embedded geometry.
The case $p=1$ has a cheap, closed-form solution.
In the other cases, $\Xi$ can be computed by solving a small convex optimization problem, smooth when $p<\infty$,
whose gradient can be evaluated with cost comparable to the SVD retraction.
This seems tractable and warrants further study.

\subsection{Bounded Rank Constraints}
\label{sec:bounded-rank}

The reader may have noticed our sleight-of-hand in the leading paragraphs:
our motivating example featured the constraint $\rank(W) \le r$,
which we tacitly tightened to $\rank(W) = r$ shortly thereafter.
Under the standard topology,
the set of matrices $W$ with $\rank(W) \le r$ does not define a manifold when $0<r<r^*$:
conceptually, it is a strata of manifolds,
and more sophisticated techniques apply (see, e.g.,~\cite{L20,LKB22,GA22}).

Working on this stratified space provides a principled way to address the geodesic incompleteness that complicated our retractions.
On the other hand, if the optimizer dynamics are consistently trending toward lower-rank solutions,
it warrants stepping back to examine the model architecture and training recipe holistically.
Are lower-rank solutions better in some sense? (E.g., generalizability?)
Or are we stumbling into an undesirable basin,
and would be better off pushing ourselves away from the singular boundary?

\subsection{Symmetry}
\label{sec:symmetry}

A common specialization of the fixed-rank manifold is to require that $W = AB^T$ is symmetric,
$W = W^T$, and, perhaps, also positive semidefinite; see, e.g.,~\cite{OHM06,MJBS09,VV10}.
In the partial isometry case,
these notions coincide: $W = UU^T$ is an orthogonal projection,
and the manifold in question can be modeled as a Grassmannian.
Indeed,
the present work spun out of an investigation into learning orthogonal projections;
that investigation continues independently.

We suggest that a symmetry constraint may be especially relevant to self-attention:
when computing the quadratic form $x^T \cdot W_{QK} \cdot x$,
there is no loss of generality to suppose $W_{QK} = W^T_{QK}$.
(This may not be relevant to cross-attention,
which employs a more general bilinear form,
$x^T \cdot W_{QK} \cdot y$.)
In the shared factor case,
each block must then have the form $W_{ij} = A L_i R_j A^T$,
where $A$ is full rank and $L_i R_j$ is invertible and, moreover, symmetric for all $i,j$,
which seems rather restrictive.
We suspect there are better ways to navigate the cost-expressivity tradeoff than factor sharing;
in particular, tensor factorizations seem like an obvious promising direction.

\subsection{Tensor Architectures}
\label{sec:tensor}

This work focused on the matrix case,
but the Riemannian approach naturally generalizes to the tensor setting; see, e.g.,~\cite{UV20} and references therein.
Our motivating application,
multihead attention,
is one of many model architectures based on rank-factored tensor parameters.
This is quite an active research area:
at the time of writing,
we observe a new rank-factored architecture proposed in the literature about every other week.
(The survey~\cite{P+21} has coverage through ${\sim}2020$.)
We selected multihead attention because it is still in widespread use after almost a decade;
separable convolutions may be the only one with longer tenure in deep learning.
Although it feels unfair to single out any particular newcomers until they've stood some test of time,
we mention that our ongoing work specifically targets
MLA~\cite{MLA},
MoLAE/LatentMoE~\cite{MoLAE,LatentMoE}, and
Mamba~\cite{Mamba}.

Another idea we wish to highlight is the tensor generalization of shared-factor architectures like group-query attention,
where sharing is replaced by more general linear relations;
see, e.g.,~\cite{K+26} for a recent instance of this idea.
Aside from improved expressivity,
perhaps such architectures will admit simpler Riemannian approaches than our grid algorithms.

\subsection{Momentum and Weight Decay}
\label{sec:momentum}

Our chosen form of momentum performs exponential smoothing on the observed Riemannian gradients,
using this weighted moving average to define the next step direction in the current tangent space.
The actual step magnitude is determined subsequently, and doesn't feed back into the momentum.
In contrast, ``heavy ball'' approaches use the actual previous step taken.
Our algorithms readily extend to the heavy ball case:
indeed, we initially expressed them as such,
but switched to the exponential smoothing approach in an effort to more closely match our experimental baseline.

We also mention weight decay,
another technique that empirically improves model accuracy.
From the Riemannian perspective,
we model conventional weight decay with a sequence of manifolds related by homothety;
the question becomes how to move our points and tangent vectors `inwards' at each optimizer step.
Our proposed approach is to scale $W$ after each optimizer step,
reinterpret $M$ as an ambient tangent vector at $W$ on the `inner' manifold,
and re-project as needed onto the submanifold tangent space.
We would like to develop a sounder theoretical foundation for this approach.

Alternatively,
some forms of weight decay can be modeled by penalizing $\|W\|$ via the objective.
Our current algorithms support this, provided the objective continues to depend on the product $W = AB^T$,
not the factors $A$ and $B$ individually.
We believe we can also support penalizing the singular values of $W$,
provided the singular vectors only influence the loss via $W$,
but have only worked through the details in the singleton fixed-rank embedded geometry.

\subsection{Other Parameterizations and Geometries}
\label{sec:parameterizations}

We have chosen to work within the AB-representation $W = AB^T$ to minimize our impact on existing model architectures.
(Reasonable minds may disagree whether our UV-representations, in the partial isometry cases, violated this goal.)
Other parameterizations may have certain advantages,
as well as implications for the choice of geometry.
We mention a few that have been promoted in the literature.
Vandereycken~\cite{V13} popularized a particular tangent space coordinate system,
which is convenient to work with when $W$ is factored as a singular value decomposition, $W = U\Sigma V^T$,
especially so when using the truncated SVD retraction.
Manopt~\cite{Manopt} relaxed this by allowing $\Sigma$ to be any invertible diagonal matrix.
A different relaxation allows $\Sigma$ to be any symmetric positive definite matrix~\cite{MMBS14}.
A factorization $W = AV^T$ or $W=U B^T$),
where $A,B$ are full rank and $U,V$ are column-orthonormal,
also has interesting theoretical benefits:
for example, it admits a metric where the exponential map is known to have a closed-form solution~\cite{AAM14}.

As we mentioned in \cref{sec:quotient},
our fixed-rank quotient metric was selected based on perceived simplicity.
A closely related metric, credited to~\cite{HM96}, was explored in~\cite{AAM14,MMBS14}.
In hindsight,
we aren't convinced our choice was substantially simpler,
and think this other quotient geometry deserves consideration.
(The distinction disappears in the partial isometry quotient case.)

\subsection{Separable Losses}
\label{sec:separable-loss}

As stated in \cref{sec:recover-GW},
we have assumed throughout that our objective $f$ depends only on $W = AB^T$,
not the factors $A$ and $B$ separately,
as this suffices for the Euclidean gradient $G_W$ to exist,
and to be recoverable from $G_A$ and $G_B$.
For example, we allow $f$ to incorporate a regularization term like $\| W \|_F^2$,
but not one like $\|A\|_F^2 + \|B\|_F^2$.
This assumption also constrains the architecture:
for example,
augmenting multihead attention with
RoPE~\cite{RoPE} or QK-norm~\cite{QKNorm}
causes $f$ to depend on $W_Q$ and $W_K$ in a more complicated manner than through the product $W_Q W_K^T$.
More subtly,
even if the assumption holds in exact arithmetic,
it may fail in finite arithmetic.

Practically speaking,
the correctness of each of our algorithms depends on the back-propagated gradients satisfying certain consistency conditions.
In the fixed-rank cases,
this condition is
$A^T G_A = G_B^T B$ (singleton)
or $\cA^T G_\cA = G_\cB^T \cB$ (grid).
In the partial isometry cases,
weaker conditions suffice,
$\Skew(U^T G_U) = \Skew(G_V^T V)$ (singleton)
or $\Skew(\cU^T G_\cU) = \Skew(G_\cV^T \cV)$ (grid),
because the symmetric components are removed by the tangent-space projection.
An objective $f$ that depends separately on the factors can still satisfy these conditions.
In our implementations,
we have considered several heuristics to reconcile violations of these conditions;
see \cref{sec:impl-recover-gradient}.
An interesting question for future work is to see if such mitigations
can extend our optimizers to support architectures that are ``close to'' low-rank,
like RoPE and QK-norm.

Another interesting direction is that if we parameterize $W$ by its singular value decomposition, $W = U\Sigma V^T$,
we can allow the loss to depend on $U,\Sigma,V$ separately and reconstruct the Riemannian gradient from $G_U$, $G_\Sigma$, and $G_V$,
provided $G_U$ and $G_V$ satisfy a consistency condition on the singular subspaces associated with repeated singular values.
This condition is automatically satisfied, for example, if the loss depends only on $W$ and $\Sigma$, but not $U$ and $V$ separately.
(This all can be deduced from a careful inspection of \cite{T16}.)
This could prove useful for supporting singular-value-based regularization.

\subsection{Retraction and Vector Transport Accuracy}
\label{sec:retractions}

The definitions of retractions and vector transports tolerate misbehavior away from the zero section of the tangent bundle.
This is not a problem for conventional gradient descent,
which invariably incorporates some form of backtracking line search.
However,
in modern deep learning applications,
most practitioners eschew line search in favor of (static) learning rate schedules.
Assuming a certain degree of regularity,
there exist learning rate schedules that guarantee convergence to an $\epsilon$-stationary point;
see, e.g.,~\cite{KR23}.
However,
it seems to be open how this theory extends to the Riemannian setting,
especially regarding the possibility of poorly behaving retractions and vector transports.
In any event, we believe there is real value in approximating the exponential map and parallel transport more accurately.

We leave for future work
a careful analysis of approximation accuracy in our algorithms,
and consideration of other retractions.
For example,~\cite{AO15} present a number of retractions for the fixed-rank embedded geometry,
including the one we used (from \cite{V13}),
and the one in \cite{SWC10}.
And in the case of Stiefel manifolds,
retractions based on the Cayley transform are popular~\cite{LLT20}.
In the case of the grid partial isometry manifold with embedded geometry (\cref{sec:grid-embedded-ortho}),
due to our hybrid approach,
it is unclear what effect increasing fidelity to the quotient geometry will have.

Taking a step back, we note that retraction is not the only principled way
to approximately traverse a manifold;
retraction-free approaches like~\cite{XGA26} are a particularly appealing alternative.

\section{Conclusion}
\label{sec:concl}

Transformer-based models are typically trained by updating the $W_{*,h}$ ($*=Q,K,V,O$) attention weights independently.
In this work we advocate treating them pairwise,
as updates to matrix parameters constrained to fixed-rank manifolds.
This is an instance of a more general technique applicable to rank-factored tensor parameters.

After tuning learning rates,
our algorithms did not consistently outperform an AdamW baseline in small-scale experiments.
In particular, the choice of geometry, momentum, and normalization did not show much of a difference.
We remain optimistic that there is merit in the Riemannian approach,
and it will become evident at larger scales.

{\small
\section*{Acknowledgements}
Charbel Sakr first posed to me the problem of learning orthogonal projections $W = UU^T$;
I thank him for our discussions ---
past, present and future ---
on solution methods for this problem.
Thanks also to Mikail Khona for keeping me abreast of the cutting edge of deep learning optimizer research,
and for his enthusiastic support for taking long shots like this one.
Thanks to Chris Baker for alerting me to~\cite{XGA26}.
Lastly, I appreciate Hao Wu's software-architectural guidance.
}

\printbibliography

\appendix

\begin{refsection}
\section{Implementation Details}
\label{sec:impl}

We give more detailed descriptions of our proposed algorithms.
Our PyTorch implementations,
available at \REPO,
closely follow the notation used in this section.
A discussion of numerical concerns is postponed to \cref{sec:numerical}.

Since we were working in Python,
we consulted the source code of Pymanopt~\cite{Pymanopt},
relevant mainly to the fixed-rank embedded geometry (\cref{sec:impl-embedded}).
Regrettably,
only late in the writing process did we think to examine the Matlab-based Manopt~\cite{Manopt},
which appears to have broader fixed-rank functionality;
we plan to review Manopt carefully in our ongoing work.

\subsection{Fixed-Rank Manifold, Embedded Geometry}
\label{sec:impl-embedded}

The following implementation was inspired by the discussion in \cite[\S7.5]{B23},
which credits \cite{V13}.
We also consulted the implementation in Pymanopt~\cite{Pymanopt}.
Our main deviations are related to our treatment of normalization and momentum,
and our insistence on keeping $W$ in AB-representation to avoid disrupting the forward/backward passes.

Our implementation involves two different factored representations of each $W \in \reals^{m\times n}_r$,
as well as two different coordinate systems on $\cT_W\reals^{m \times n}_r$.

For each $W \in \reals^{m\times n}_r$, we consider two factored representations.
\begin{itemize}

\item An \emph{AB-representation} for $W$ is a pair $(A,B) \in \reals^{m\times r}_* \times \reals^{n\times r}_*$ such that $W = AB^T$.
Two such pairs $(A,B),(A',B')$ are equivalent if $AB^T = A'{B'}^T$.
This is an equivalence relation;
each equivalence class equals $\{(AS, BS^{-T}) : S \in \reals^{r\times r}_*\}$ for any representative $(A,B)$.

\item A \emph{USV-representation} for $W$ is a triple $(U,S,V) \in \St_{m,r} \times \reals^{r\times r}_* \times \St_{n,r}$ such that $W = USV^T$.
Two such triples $(U,S,V),(U',S',V')$ are equivalent if $USV^T = U'S'{V'}^T$.
This is an equivalence relation;
each equivalence class equals $\{(UQ_U, Q_U^TSQ_V, VQ^T_V) : Q_U,Q_V \in \St_{r,r}\}$ for any representative $(U,S,V)$.
\end{itemize}

For each fixed USV-representation of $W$,
we define two coordinate systems on $\cT_W(\reals^{m \times n}_r)$:
\begin{itemize}
\item \emph{XKY-coordinates} are the triples $(X,K,Y) \in \reals^{m\times r} \times \reals^{r\times r} \times \reals^{n \times r}$ such that $U^T X = V^T Y = 0_{r\times r}$.
This set of triples is a vector space
(defining operations componentwise),
and is isomorphic to $\cT_W(\reals^{m \times n}_r) \ni \Xi$
via $\Xi = UKV^T + XV^T + UY^T$,
justifying the terminology.
(Vandereycken used the notation $(U_p,M,V_p)$,
but we prefer to reserve `$U$' and `$V$' for column-orthonormal matrices,
and we've already used `$M$' for momentum.)

\item The \emph{CD-coordinates} are the pairs $(C,D) \in \reals^{m\times r} \times \reals^{n \times r}$ such that $C = \Xi V$ and $D =\Xi^T U$ for some $\Xi \in \cT_W(\reals^{m \times n}_r)$.
This linear relation is invertible, with
$\Xi = CV^T+UD^T-UU^T C V^T = CV^T + UD^T-UD^T VV^T$,
again justifying the terminology.
Given XKY-coordinates $(X,K,Y)$,
we have $C = UK + X$ and $D = VK^T + Y$;
going the other direction,
$K = U^TC = D^TV$,
$X = (I - UU^T)C$, and
$Y = (I - VV^T)D$.
\end{itemize}

Let $W \in \reals^{m \times n}_r$ be our current iterate,
with AB-representation $(A,B)$ and Euclidean factor gradients $G_A$ and $G_B$.

\begin{itemize}
\item (Step 0) Compute (thin) QR factorizations,
$A=Q_AR_A$ and $B=Q_BR_B$, or reuse existing ones if available.
The triple $(U,S,V) = (Q_A, R_AR_B^T,Q_B)$ is an USV-representation of the current iterate $W$.
As we will see, this step can be amortized.

\item (Steps 1 and 2) An ambient gradient $G_W$ satisfies
$G_A = G_W B = G_W V R_B$ and $G_B = G_W^T A = G^T_W U R_A$.
The Riemannian gradient is $\Pi_W(G_W)$,
where $\Pi_W(Z) = P_UZ + ZP_V - P_UZP_V$, with $P_U = UU^T$ and $P_V = VV^T$.
Note the following trick: $\Pi_W(Z)V = ZV$ and $U^T \Pi_W(Z) = U^T Z$.
Thus, sending $G_W$ through $\Pi_W$ and applying this trick,
we readily obtain $G^R_W$ in CD-coordinates:
$C_G = G^R_W V = G_W V = G_A R_B^{-1}$ and
$D_G = (G^R_W)^T U = G^T_W U = G_B R_A^{-1}$.
In summary, these steps boil down to two triangular solves:
\[
(C_G, D_G) = ( G_A R_B^{-1}, G_B R_A^{-1})
\text.
\]

\item (Step 3)
Supposing we already have momentum $M$ in CD-coordinates, $(C_M, D_M)$,
update it by
\[ (C_M, D_M) \gets \nu (C_M, D_M) + (1-\nu) (C_G,D_G)
\text. \]
Our next step $T$ is the momentum, normalized and scaled by the negative learning rate:
$T = \gamma M$,
where $\gamma=-\eta/\mu$,
$\eta$ is the learning rate,
and $\mu = \| M \|$
is what remains to be computed.
It turns out we can compute $\mu$ more efficiently by switching to XKY-coordinates.
Even if normalization is not desired, we still perform this conversion,
because it cheapens the following SVD retraction (Step~4) as well.

Let's convert $(C_M,D_M)$ to XKY-coordinates $(X,K,Y)$.
First, recover $K = U^T C_M = D_M^TV$;
in exact arithmetic, we need only compute one of these expressions,
and we'd decide based on the relative sizes of $m$ and $n$.
As a numerical safeguard, we compute both and take the average,
\[
K = \frac12(U^T C_M + D_M^TV)
\text.
\]
The computations of $X$ and $Y$ can reuse $K$:
\begin{align*}
X &= (I-UU^T)C_M = C_M-UK
\\
Y &= (I-VV^T)D_M=D_M-VK^T
\text.
\end{align*}
Introducing QR factorizations
$X = Q_X R_X$ and $Y = Q_Y R_Y$,
we see that our unnormalized momentum is
\[
\begin{bmatrix} U & Q_X \end{bmatrix}
\begin{bmatrix} K & R_Y^T \\ R_X & 0\end{bmatrix}
\begin{bmatrix} V & Q_Y \end{bmatrix}^T
\text.
\]
Recall we target both the Frobenius and spectral norms,
and recall both are orthogonally invariant.
Thus, in both cases, $\mu$ equals the norm of the middle $2r$-by-$2r$ matrix.
Compute $\mu$ now, giving $\gamma$.
By linearity, $(\gamma X,\gamma M,\gamma Y)$ give XKY-coordinates for $T$.

\item (Step 4)
Now we turn to our SVD retraction, which is a truncated SVD of $W+T$.
Given $W$ in USV-coordinates from Step~0,
and $T$ in XKY-coordinates from Step~3,
this matrix sum equals
\[
W+T=
\begin{bmatrix} U & Q_X \end{bmatrix}
\begin{bmatrix} S + \gamma K & \gamma R_Y^T \\ \gamma R_X & 0\end{bmatrix}
\begin{bmatrix} V & Q_Y \end{bmatrix}^T
\text.
\]
Similarly to the norm computation in Step~3,
since $[U,Q_X]$ and $[V,Q_Y]$ both have orthonormal columns,
it suffices to perform our truncated SVD on the inner $2r$-by-$2r$ matrix.
Let $(U_r, \Sigma_r,V_r)$ denote the (generally non-unique) result.
Compute our next iterates via their QR factorizations,
\[
\hat A = \hat U \hat R_A
\qquad\text{and}\qquad
\hat B = \hat V \hat R_B
\text,\]
where
\[
\hat U = [U,Q_X]U_r
\text,\quad
\hat V = [V,Q_Y]V_r
\text,\quad\text{and}\quad
\hat R_A = \hat R_B = \Sigma^{1/2}_r
\]
which can be saved for the next iteration.
(Note that the R-factors are now diagonal, even if they weren't before.)

\item (Step 5)
We transport our momentum $M$ to $\cT_{\hat W}$
by interpreting it as an ambient tangent vector at $\hat W$,
then projecting it, i.e., $\hat M = \Pi_{\hat W}(M)$.
Working within CD-coordinates,
we obtain
\begin{align*}
C_{\hat M} &= C_M V_{r,1} + U\big(R_Y V_{r,2}\big)
\\
D_{\hat M} &= D_M U_{r,1} + V\big(R_X U_{r,2}\big)
\text,
\end{align*}
where the grouping of the right-hand terms suggests how to minimize arithmetic,
and where we've split $U_r$ and $V_r$ into $r$-by-$r$ submatrices,
\[
U_r = \begin{bmatrix} U_{r,1} \\ U_{r,2} \end{bmatrix}
\qquad\text{and}\qquad
V_r = \begin{bmatrix} V_{r,1} \\ V_{r,2} \end{bmatrix}
\text.
\]
To derive this, we first use the same trick we used with $G^R_W$.
That is, starting with the C-coordinate,
\[
C_{\hat M} = M \hat V = (M V)V_{r,1}+(M Q_Y)V_{r,2}
\text.
\]
The first parenthesized term is just $C_M$.
The second can be shown to equal $UR_Y$:
expand $M$ in terms of $C_M$ and $D_M$,
postmultiply by $Q_Y$,
and substitute $V^TQ_Y = 0$.
The D-coordinate is symmetric:
\[
\hat D_M = M^T \hat U = (M^TU)U_{r,1}+(M^TQ_X)U_{r,2}
\text,
\]
then substitute $M^TU=D_M$ and $M^TQ_X = VR_X$,
the latter from expanding $M^T$ in CD-coordinates,
postmultiplying by $Q_X$,
and substituting $U^TQ_X = 0$.

\end{itemize}

In our initial experiments with this implementation,
we observed that the saved Q-factors lost orthogonality due to floating-point error in the updates $\hat U$ and $\hat V$ in Step~4,
leading to training divergence.
We identified four different ways of performing reorthogonalization;
we prefer the following one for its simplicity:

\begin{itemize}
\item (Step~4, reorthogonalization):
After computing $\hat A$ and $\hat B$,
compute explicit QR factorizations
$\hat A = \hat U \hat R_A$ and $\hat B = \hat V \hat R_B$,
overwriting the old ones.

\item (Step~5, reorthogonalization):
Evaluate the same formulas,
except substitute
\[
\begin{bmatrix} U_{r,1} \\ U_{r,2} \end{bmatrix} = \begin{bmatrix} U^T \hat U \\ Q_X^T \hat U \end{bmatrix}
\qquad\text{and}\qquad
\begin{bmatrix} V_{r,1} \\ V_{r,2} \end{bmatrix} = \begin{bmatrix} V^T \hat V \\ Q_Y^T \hat V \end{bmatrix}
\text.
\]
\end{itemize}

\subsubsection{Grid Case}
\label{sec:impl-grid-embedded}

To handle shared weights,
we simply concatenate the $A$ and $B$ factors,
and associated gradients $G_A$ and $G_B$,
and apply the preceding algorithm.

As promised in \cref{sec:grid-embedded},
we have a stronger sufficient condition
for the retraction to be well defined,
even though our implementation doesn't actually use either condition.

Continuing notation from \cref{sec:grid-embedded},
let's additionally abbreviate $\delta = \|\Xi\|_2$,
and suppose that $\delta < S$,
justifying it later.
Recall our task is to derive an upper bound on $\|R - \cW\|_2$.
By the triangle inequality and Eckart-Young-Mirsky,
recalling $R$ is optimal,
\[
\|R - \cW\|_2
\le
\delta + \| (\cW + \Xi) - R\|_2
\le
\delta + \| (\cW + \Xi) - \hat R\|_2
\]
for any conformable rank-$r$ matrix $\hat R$;
we will construct a specific such $\hat R$.
Take any SVD, $\cW = U\Sigma V^T$,
and any conformable $U_\bot,V_\bot$ such that $[U,U_\bot]$ and $[V,V_\bot]$ are orthogonal.
There exist conformable $E,K,F$ such that
\[ \Xi =
\begin{bmatrix} U & U_\bot \end{bmatrix}
\begin{bmatrix} K & F^T \\ E & 0 \end{bmatrix}
\begin{bmatrix} V^T \\ V^T_\bot \end{bmatrix}
\text.
\]
(In terms of XKY-coordinates, $X=U_\bot E$ and $Y = V_\bot F$.)
Note that $\Sigma + K$ is invertible:
$\sigma_r(\Sigma+K) \ge S - \|K\|_2 \ge S - \delta > 0$,
again via Weyl and Cauchy,
plus orthogonal invariance of $\|\cdot \|_2$,
and our assumption that $\delta < S$.
This allows us to define the rank-$r$ matrix
\[ \hat R =
\begin{bmatrix}
U &
U_\bot
\end{bmatrix}
\begin{bmatrix} I \\ E(\Sigma + K)^{-1} \end{bmatrix}
(\Sigma + K)
\begin{bmatrix} I & (\Sigma + K)^{-1} F^T \end{bmatrix}
\begin{bmatrix} V^T \\ V^T_\bot \end{bmatrix}
\text,
\]
chosen so that its associated approximation error is isolated to the trailing block,
\[
\begin{bmatrix} U^T \\ U^T_\bot \end{bmatrix}
((\cW + \Xi) - \hat R)
\begin{bmatrix} V & V_\bot \end{bmatrix}
=
\begin{bmatrix}
0 & 0 \\ 0 & -E(\Sigma + K)^{-1} F^T
\end{bmatrix}
\text.
\]
Continuing our chain of inequalities,
\[
\|R - \cW\|_2 - \delta
\le
\| (\cW + \Xi) - \hat R\|_2
=
\|E (\Sigma + K)^{-1} F^T \|_2
\le
\|E\|_2 \|(\Sigma+K)^{-1}\|_2 \|F^T\|_2
=
\frac{\|E\|_2 \|F\|_2}{\sigma_r(\Sigma+K)}
\le
\frac{\delta^2}{S - \delta}
\text,
\]
again appealing to Weyl, Cauchy, and orthogonal invarience.
Thus
$\|R - \cW\|_2 \le \delta + \delta^2 / (S-\delta) = S\delta/(S-\delta)$;
solving $S\delta/(S-\delta) < s$ for $\delta$ reveals the condition claimed in \cref{sec:grid-embedded}.
Finally, note that $s \le S$ (Cauchy again),
so $\delta < sS / (s + S) \le S/2$,
thus there was no loss of generality earlier to suppose $\delta < S$.

The simpler but more conservative condition given earlier,
$\|\Xi\|_2 < s / 2$,
is a corollary of this stronger result: $sS / (s + S) \ge s/2$ since $s \le S$.
However, the earlier reasoning did not require that $\Xi$ is a tangent vector,
so perhaps that condition can be more robust to numerical issues,
like those discussed in \cref{sec:impl-recover-gradient,sec:momentum-decoherence}.

We note that the same $\hat R$ can be used in a similar argument for the singleton partial isometry embedded case to obtain the condition
$\|E(I+K)^{-1}F^T\|_2 < \sigma_r(I+K)$,
which implies the one we gave in \cref{sec:embedded-ortho},
$\|\Xi\|_2 < 1/2$.
The main change in the argumentation is that $\Sigma+K=I+K$ is always invertible since $K = -K^T$.

\subsection{Fixed-Rank Manifold, Quotient Geometry}
\label{sec:impl-quotient}

Let $W \in \reals^{m \times n}_r$ be our current iterate,
with AB-representation $(A,B)$ and Euclidean factor gradients $G_A$ and $G_B$.

\begin{itemize}
\item (Steps 1 and 2)
The horizontal lift of the Riemannian gradient at $AB^T \in \cM$
is
\[
\big(G^R_A, G^R_B\big)=
\big(G_A (B^TB)^{-1}, G_B (A^TA)^{-1}\big) =
\big((G_A L_B^{-T}) L_B^{-1}, (G_B L_A^{-T}) L_A^{-1}\big)
\text,
\]
where $A^TA = L_A L_A^T$ and $B^TB = L_B L_B^T$ are Cholesky factorizations,
which we compute here, or reuse from before if available.
We perform the four triangular solves in the manner indicated by grouping.

\item (Step 3)
Supposing we already have the horizontal lift of the momentum,
update it as
\[
(M_A, M_B) \gets \nu (M_A, M_B) + (1-\nu)(G^R_A, G^R_B)
\text.
\]
Our next step is $(T_A, T_B) = \gamma (M_A, M_B)$, where
$\gamma = -\eta / \mu$
with $\mu = \sqrt{\|M_A L_B\|^2_F + \|M_B L_A\|^2_F}$ if normalization is requested,
otherwise $\gamma = -\eta$.

\item (Step 4):
Our additive retraction is simply
\[ (\hat A, \hat B) = (A + T_A, B + T_B) \]
For future use,
we form Cholesky factorizations,
$\hat A^T \hat A = \hat L_A \hat L_A^T$ and $\hat B^T \hat B= \hat L_B \hat L_B^T$.

\item (Step 5):
Our vector transport is just horizontal projection.
We first compute
\begin{align*}
\Lambda
&= \frac12 \left(\left(\hat A^T \hat A\right)^{-1} \hat A^T M_A - M_B^T \hat B \left(\hat B^T \hat B\right)^{-1}\right)
\\
&= \frac12 \left(
\hat L_A^{-T} \left(\hat L_A^{-1} \left(\hat A^T M_A\right)\right)
-
\left(\left(M_B^T \hat B \right) \hat L_B^{-1} \right) \hat L_B^{-T}\right)
\end{align*}
using the Cholesky factorizations just obtained, then our updated momentum
\[
(\hat M_A, \hat M_B) = (M_A - \hat A \Lambda, M_B + \hat B \Lambda^T)
\text.
\]

\end{itemize}

\subsubsection{Grid Case}
\label{sec:impl-grid-quotient}

To handle shared weights,
we simply concatenate the $A$ and $B$ factors as appropriate,
and apply the preceding algorithm.

\subsection{Partial Isometry Manifold, Embedded Geometry}
\label{sec:impl-embedded-ortho}

The AB and USV representations on $\reals^{m \times n}_r$ coincide on $\cP^{m, n}_r$, as follows.
For each $W \in \cP^{m, n}_r$,
a \emph{UV-representation} for $W$ is a pair $(U,V) \in \St_{m,r} \times \St_{n,r}$ such that $W = UV^T$.
Two such pairs $(U,V),(U',V')$ are equivalent if $UV^T = U'{V'}^T$.
This is an equivalence relation;
each equivalence class equals $\{(UQ, VQ) : Q \in \St_{r,r}\}$ for any representative $(U,V)$.

The XKY and CD coordinates from $\cT\reals^{m \times n}_r$ simplify as well.
In particular, we have the additional constraint that $K^T = -K$.

\begin{itemize}
\item (Step 0)
Unlike the fixed-rank case, we don't need to compute QR factorizations;
in particular,
we are given a UV-representation $(U,V)$ as input.

\item (Steps 1 and 2):
An ambient gradient $G_W$ satisfies $G_U = G_W V$ and $G_V = G_W^T U$.
The Riemannian gradient is $\Pi_W(G_W)$,
where
$\Pi_W(Z) = (I-P_U)ZP_V + P_UZ(I-P_V) + U \Skew(U^T Z V) V^T$.
We go straight to CD-coordinates:
\[
(C_G, D_G) = (G_U - U K , G_V - V K )
\]
where $K = \Sym(U^T G_U) = \Sym(V^T G_V)$ in exact arithmetic,
so to minimize arithmetic we would decide based on $m$ vs.\ $n$.
As a mild numerical safeguard,
our implementation averages the two, $K = \Sym(U^T G_U + V^T G_V)/2$.

\item (Step 3)
This is essentially the same Step~(3) as in the fixed-rank embedded case;
we highlight just what changes.

Since $K = U^T C_M = D_M^TV = -K^T$ in exact arithmetic,
we modify our numerical robustness heuristic,
$K = \Skew (U^T C_M - V^T D_M) / 2$.
Now $K = -K^T$ holds in finite precision.

\item (Step 4)
The only changes here are that $S = I$,
and we don't need to scale by singular values, since they are all one.

\item (Step 5)
Following the fixed-rank embedded case,
we obtain CD-coordinates $(C_{\hat M}, D_{\hat M})$ for an element $\hat M \in \cT_{\hat W} \reals^{m\times n}_r$;
we now need to project this via $\Pi_{\hat W}$.
This works similarly to computing the Riemannian gradient;
the update is
\[
(C_{\hat M}, D_{\hat M})
\gets
(C_{\hat M} - \hat U K, D_{\hat M} - \hat V K)
\]
where $K = \Sym(\hat U^T C_{\hat M}) = \Sym(\hat V^T D_{\hat M})$ in exact arithmetic,
but again we instead take an average,
$K = \Sym(\hat U^T C_{\hat M} + \hat V^T D_{\hat M})/2$.

\end{itemize}

The reorthogonalization approach from the fixed-rank case applies here as well.
Its justification is a bit more subtle here because we are actually changing the base point, not just our basis of it.
In exact arithmetic, the SVD retraction gives a partial isometry $W_* = U_*V_*^T$,
while in finite precision we have $W = U V^T$,
where $U,V$ are close to column-orthonormal.
Explicitly orthogonalizing $U,V$ via QR or polar decomposition gives
$\hat U_\mathrm{QR}, \hat V_\mathrm{QR}$ or
$\hat U_\mathrm{polar},\hat V_\mathrm{polar}$.
Letting
$\hat W_\mathrm{QR} =\hat U_\mathrm{QR} \hat V_\mathrm{QR}^T$ and
$\hat W_\mathrm{polar} = \hat U_\mathrm{polar} \hat V_\mathrm{polar}^T$,
we have that
\begin{align*}
\| W - \hat W_\mathrm{QR} \| &= \| R_U R_V^T - I \|\text,
&
\| W - \hat W_\mathrm{polar} \| &= \| H_U H_V - I\|\text,
\end{align*}
in any orthogonally invariant norm,
where $R_U,R_V$ are the R-factors of the QR decomposition,
and $H_U,H_V$ are the H-factors of the polar decomposition.
From here, we can show that both $\hat W_\mathrm{QR}$ and $\hat W_\mathrm{polar}$ approximate $\hat W$ to first order in the orthogonality defects.
(The triangle inequality provides a bound on the total error,
$\|\hat W - W_*\| \le \| W - W_*\| + \|W - \hat W\|$.)
In the polar case, we can specifically show that
\[ \| W - \hat W_\mathrm{polar} \|_2 \le  \frac12(\delta_U + \delta_V) + O(\delta^2_U + \delta^2_V) \text,\]
where
$\delta_U = \|U^TU - I\|_2$ and $\delta_V=\|V^TV - I\|_2$.
We have been unable to find a similar bound for QR that doesn't involve an R-factor on the leading term.
For this reason, we suggest using the polar decomposition to reorthogonalize.

\subsubsection{Grid Case}
\label{sec:impl-grid-embedded-ortho}

As discussed before,
the grid partial isometry ``embedded'' algorithm ended up hybridizing with the quotient geometry:
halfway through the algorithm we lift to a product space, using the horizontality condition
\[ J \sum_i U_i^T \Xi_{U_i} + I \sum_j V_j^T \Xi_{V_j} = 0 \]
from the grid partial isometry quotient geometry (which happens to be the same in the canonical case),
treated in the following section.
We keep momentum upstairs, form our step upstairs, and perform retractions and vector transports upstairs.
However, we will work downstairs briefly to derive the Riemannian gradient,
and to normalize our momentum.
We will need to convert between grid XKY coordinates downstairs and horizontal vectors upstairs.

In the grid case, the XKY coordinates for $\Xi \in \cT_{(U_iV_j)_{ij}} \cG$ can be written blockwise:
for each $i,j$,
\[
\Xi_{ij} = U_i K_{ij} V_j^T + X_i V_j^T + U_i Y_j^T
\text,
\]
satisfying the same constraints as the singleton partial isometry case,
$K_{ij} = -K^T_{ij}$ and $U_i^T X_i = V_j^T Y_j = 0$,
plus an additional condition,
that $K_{ij} + K_{i'j'} = K_{i'j} + K_{ij'}$ for all $i',j'$.
The conditions on $(K_{ij})_{ij}$ can be equivalently stated as saying there exist skew-symmetric matrices $(\Omega_i)_i$ and $(\Psi_j)_j$ such that each $K_{ij} = \Omega_i - \Psi_j$ for all $i,j$.
In particular, we can represent the $IJ$ matrices $(K_{ij})_{ij}$ in terms of the $I+J$ matrices $(\Omega_i)_i$ and $(\Psi_j)_j$.
In principle, this representation is not unique:
for any skew-symmetric $\Lambda$, we can take $\Omega_i + \Lambda$ and $\Psi_j + \Lambda$.
We will pick $\Omega_i,\Psi_j$ to satisfy $J \sum_i \Omega_i + I \sum_j \Psi_j = 0$;
this imposes the (quotient geometry) horizontality condition when lifting to factor space.

Given a tangent vector $((\Xi_{U_i})_i,(\Xi_{V_j})_j)$ in factor space,
we can reconstruct grid-XKY coordinates as follows:
\begin{equation*}
\begin{aligned}
\Omega_i &= U_i^T \Xi_{U_i}
&
\Psi_j &= V_j^T \Xi_{V_j}
\\
X_i &= \Xi_{U_i} - U_i \Omega_i
&
Y_j &= \Xi_{V_j} - V_j \Psi_j
\end{aligned}
\end{equation*}
And given grid-XKY coordinates, we can lift it to factor space,
\[
\Xi_{U_i} = X_i + U_i \Omega_i
\qquad
\Xi_{V_j} = Y_j + V_j \Psi_j
\text;
\]
in exact arithmetic, this is not only tangent, but also horizontal.

\begin{itemize}

\item (Steps 1 and 2)
Compute the following quantities:
\begin{gather*}
\begin{aligned}
\Alpha_i &= \Skew(U_i^T G_{U_i})\text,
&
\Beta_j &= \Skew(G_{V_j}^T V_j)\text,
\\
X_i &= \frac{1}{J} (I-U_i U_i^T)G_{U_i} \text,
&
Y_j &= \frac{1}{I}(I-V_j V_j^T)G_{V_j}\text,
\end{aligned}
\\
\Gamma = \frac12\Big(\sum_i \Alpha_i + \sum_j \Beta_j \Big)\text,
\\
\begin{aligned}
\Omega_i &= \frac{1}{J}\Alpha_i - \frac{1}{2IJ}\Gamma
\text,
&
\Psi_j &= \frac{1}{2IJ}\Gamma - \frac{1}{I}\Beta_j
\text.
\end{aligned}
\end{gather*}
Here,
$\Gamma$ averages two quantities that should be equal in exact arithmetic,
as a numerical safeguard.
(We applied the horizontality condition here to resolve the nonuniqueness in choosing $\Omega_i$ and $\Psi_j$.)

Having obtained grid-XKY coordinates for the embedded Riemannian gradient,
we lift to factor space,
\begin{align*}
G^R_{U_i} &= X_i + U_i \Omega_i
\\
G^R_{V_j} &= Y_j + V_j \Psi_j
\text.
\end{align*}

\item (Step 3)
The rest of the steps proceed as in the grid partial isometry quotient/canonical geometries, with the exception of the norm computation.
Our updated momentum $M$ is represented upstairs by $((M_{U_i})_i,(M_{V_j})_j)$.
Directly computing $\| M \|^2_F = \sum_{ij} \| M_{ij} \|^2_F$ requires generating $IJ$ matrices of size $m_i$-by-$n_j$.
By converting to grid-XKY coordinates, we can reduce this to $2(I+J)$ matrices, of dimensions $m_i \times r$, $n_j \times r$, and $r \times r$ (but not $m_i \times n_j$),
and evaluate our metric norm blockwise as
\[
\| M_{ij} \|_F^2
=
\|\hat K_{ij} \|_F^2 + \| \hat X_i \|_F^2 + \| \hat Y_j \|_F^2
\text,
\]
introducing hats to distinguish from the earlier grid-XKY coordinates.
In detail,
\begin{align*}
\hat \Omega_i &= U_i ^T M_{U_i}
&
\hat \Psi_j  &= V_j^T M_{V_j}
\\
\hat X_i &= M_{U_i} - U_i \hat \Omega_i
&
\hat Y_j &= M_{V_j} - V_j \hat \Psi_j
\end{align*}
giving
\[
\| M \|^2_F =
J \sum_i \Big( \| \hat X_i \|_F^2 + \| \hat \Omega_i\|_F^2\Big)
+
I \sum_j \Big( \| \hat Y_j \|_F^2 + \| \hat \Psi_j \|_F^2\Big)
-
2\left\langle \sum_i \hat \Omega_i, \sum_j \hat \Psi_j \right\rangle
\]
In the absence of momentum ($\nu = 0$),
we can reuse the Riemannian gradient's XKY coordinates here.

Another approach, that enables us to compute any Schatten-$p$ norm,
is as follows.
Let $\cU,\cV,\cM_\cU,\cM_\cV$ denote vertically stacked $(U_i)_i,(V_j)_j,(M_{U_i})_i, (M_{V_j})_j$.
Assembling blocks over the whole grid,
the tangent matrix is $M = \cM_\cU \cV^T + \cU \cM_\cV^T = \cA \cB^T$,
where $\cA = [\cM_\cU, \cU]$ and $\cB = [\cV, \cM_\cV]$ (note the swap).
The nonzero singular values of $M$ equal the square roots of the nonzero eigenvalues of $(\cA^T\cA)(\cB^T \cB)$.
(Despite being a nonsymmetric matrix, its eigenvalues are all real and nonnegative, in exact arithmetic.)
In particular, we have
\begin{align*}
G_\cA = \cA^T\cA &=
\begin{bmatrix}
\sum_i X_i^T X_i + \Omega_i^T \Omega_i & \sum_i \Omega_i^T \\ \sum_i \Omega_i & I \cdot I_{r \times r}
\end{bmatrix}
\\
G_\cB = \cB^T\cB &=
\begin{bmatrix}
J \cdot I_{r \times r} & \sum_j \Psi_j
\\
\sum_j \Psi_j^T & \sum_j Y_j^T Y_j + \Psi_j^T \Psi_j
\end{bmatrix}
\end{align*}
The spectral norm is the square root of the maximum eigenvalue of $G_\cA G_\cB$.
Rather than invoke a nonsymmetric eigensolver,
we leverage the fact that
the nonzero eigenvalues of $G_\cA G_\cB$ equal those of $S G_\cB S^T$ for any factorization $G_\cA = S^TS$.
(Alternatively, a splitting of $G_\cB$ can be used.)
For robustness we take $S = Q \Lambda^{1/2}$ from a spectral decomposition $G_\cA = Q \Lambda Q^T$,
and leave investigation of faster (e.g., Cholesky-based) alternatives for future work.
(This also yields an alternative algorithm for the metric norm,
$\| M \|_F^2 = \Tr ( G_\cA G_\cB)$,
although the earlier one seems much more efficient.)

\item (Steps 4 and 5)
Here we exactly follow the grid partial isometry quotient case, discussed shortly.

\end{itemize}

In the case $I=J=1$, the grid algorithm (implicitly) constructs the same downstairs Riemannian gradient as the singleton algorithm,
as well as the same step when $\nu = 0$ or $M = 0$ (e.g., on the first iteration).
However, the retraction and vector transport are fundamentally different,
so we don't expect the same sequence of iterates.

\subsection{Partial Isometry Manifold, Quotient Geometry}
\label{sec:impl-quotient-ortho}

\begin{itemize}

\item (Steps 1 and 2)
Our lifted Riemannian gradient is $(G^R_U, G^R_V) = (G_U - U K, G_V - VK)$,
where $K = \Sym( G_U^T U) = \Sym(G_V^T V)$ in exact arithmetic.
Once again, we compute $K$ as an average, $K = \Sym( G_U^T U + G_V^T V) / 2$.

\item (Step 3)
Our norm is $\mu = \sqrt{\| M_U\|^2_F + \| M_V \|_F^2}$.

\item (Step 4)
We implement the QR retraction via a (thin) Householder QR, with postprocessing to `canonicalize' the R-factor.
We implement the polar retraction via SVD.
There are lots of interesting ways to accelerate these operations, which we do not explore here.

\item (Step 5)
The most straightforward approach is to first project the momentum onto the tangent space at the retracted point,
\begin{align*}
M'_{\hat U} &= M_U - \hat U \Sym( M_U^T \hat U)
\\
M'_{\hat V} &= M_V - \hat V \Sym( M_V^T \hat V)
\text,
\end{align*}
then remove the vertical component,
\begin{align*}
M_{\hat U} &= M'_{\hat U} - \hat U \Omega
\\
M_{\hat V} &= M'_{\hat V} - \hat V \Omega
\text,
\end{align*}
where
\[
\Omega = \frac12 (\hat U^T M'_{\hat U} + \hat V^T M'_{\hat V})
\text.
\]
It is possible to considerably reduce the arithmetic involved in matrix multiplications, e.g.,
\begin{align*}
M_{\hat U} &= M_U - \hat U (\hat U^T M_U + \Skew(\hat V^T M_V)/2)
\\
M_{\hat V} &= M_V - \hat V (\hat V^T M_V + \Skew(\hat U^T M_U)/2)
\text.
\end{align*}
We suspect this may yield slightly worse accuracy than the straightforward approach,
so we deferred investigation of this optimization,
along with all of the other $K$-related optimizations we've mentioned.

\end{itemize}

\subsubsection{Grid Case}
\label{sec:impl-grid-quotient-ortho}

\begin{itemize}

\item (Steps 1 and 2)
Our lifted Riemannian gradients are the same as before, except computed factorwise, and weighted:
\begin{align*}
G^R_{U_i} &= \frac{1}{J}\Big(G_{U_i} - U_i \Sym(G_{U_i}^T U_i)\Big)
\\
G^R_{V_j} &= \frac{1}{I}\Big(G_{V_j} - V_j \Sym(G_{V_j}^T V_j)\Big)
\end{align*}
We did not explore a generalization of the averaging technique from the singleton case;
we suggest a more principled approach, discussed in \cref{sec:impl-recover-gradient}.%

\item (Step 3)
Our norm is $\mu = \sqrt{J \sum_i \| M_{U_i}\|^2_F + I\sum_j\| M_{V_j} \|_F^2}$.

\item (Step 4)
Retraction is the same, just factorwise.

\item (Step 5)
Vector transport is the same, just factorwise, with one change:
\[
\Omega = \frac{J \sum_i \hat U_i^T M'_{\hat U_i} + I \sum_j \hat V_j^T M'_{\hat V_j}}{2IJ}
\]
The optimization we mentioned before extends here as well.

\end{itemize}

\subsection{Partial Isometry Manifold, Canonical Geometry}
\label{sec:impl-canonical}

Due to considerable overlap of logic,
we implement the partial isometry quotient and canonical geometries in the same routine.
The only differences between the two are in the gradient and norm computations.

\begin{itemize}

\item (Steps 1 and 2)
Our lifted Riemannian gradient is $(G^R_U, G^R_V) = (G_U - U K, G_V - VK^T)$,
where $K = G_U^T U = V^T G_V$ in exact arithmetic.
We compute $K$ as an average, $K = (G_U^T U + V^T G_V) / 2$.

\item (Step 3)
Our norm is $\mu = \sqrt{\| M_U\|^2_F - \frac12\|U^T M_U\|_F^2 + \| M_V \|_F^2 - \frac12\|V^T M_V\|_F^2 }$.

\item (Step 4)
We reuse the retractions from the partial isometry quotient geometry.

\item (Step 5)
The horizontal space in the canonical geometry turns out to be the same as in the partial isometry quotient geometry,
so our vector transport is identical to the quotient case.

\end{itemize}

\subsubsection{Grid Case}
\label{sec:impl-grid-canonical}

\begin{itemize}

\item (Steps 1 and 2)
Our lifted Riemannian gradients are the same as in the singleton case, except computed factorwise, and weighted:
\begin{align*}
G^R_{U_i} &= \frac{1}{J}\Big(G_{U_i} - U_i (G_{U_i}^T U_i)\Big)
\\
G^R_{V_j} &= \frac{1}{I}\Big(G_{V_j} - V_j (G_{V_j}^T V_j)\Big)
\end{align*}
Again,
we did not explore a generalization of the averaging technique from the singleton case,
and instead suggest a more principled approach (\cref{sec:impl-recover-gradient}).

\item (Step 3)
Our norm is $\mu = \sqrt{J\sum_i (\| M_{U_i}\|^2_F - \frac12\|U_i^T M_{U_i}\|_F^2) + I \sum_j( \| M_{V_j} \|_F^2 - \frac12\|V_j^T M_{V_j}\|_F^2) }$.

\item (Steps 4 and 5)
Retraction and vector transport are identical to the grid partial isometry quotient geometry.

\end{itemize}

\section{Numerical Details}
\label{sec:numerical}

In this section we explore several ways in which things can go awry.
Our ongoing work addresses diagnostics for these failure modes,
as well as mitigations.

\subsection{Rank Deficiency}

Simply put,
if a retraction returns $W$ with rank less than $r$,
the results of the optimizer step cannot be trusted.
This problem is specific to the fixed-rank cases.
It's tempting to blame such failures on the geodesic incompleteness of the fixed-rank manifold,
but the situation is murky since we are using retractions and vector transports,
not necessarily following true geodesics.
Whatever the underlying reason,
the error is fatal:
in both embedded and quotient geometries,
attempting to take another step will fail spectacularly almost immediately,
by attempting to invert a singular matrix.

Currently we implement two cheap, heuristic safeguards.
In the case of the fixed-rank embedded geometry,
we currently implement a cheap, heuristic safeguard:
our SVD retraction saturates vanishing singular values to a small positive constant.
In the case of the fixed-rank quotient geometry,
if we detect a failure from the Cholesky driver,
we currently terminate the training.
(We didn't observe this in our experiments.)
As a cheap safeguard,
we propose to perturb the failing problems (Gram matrices) by adding a small positive constant to their diagonals, then repeat.

\subsection{Loss of orthogonality}

Our fixed-rank embedded algorithm maintains QR factorizations $A = U R_A$ and $B = V R_B$,
stored as optimizer states.
(In the grid case, $A$ and $B$ are stacked matrices.)
We expect $A = U R_A$ and $B = VR_B$ to hold with close to machine precision at any iteration.
However,
without explicit reorthogonalization,
$U^TU$ and $V^TV$ may drift further from $I$ over time.
Similarly, in our singleton partial isometry embedded algorithm,
the model weights  $U$ and $V$ will drift away from their Stiefel manifolds.
We have observed such losses of orthogonality to have catastrophic effects on training,
thus our implementations take a conservative approach and reorthogonalize every iteration.

There are several different ways to incorporate reorthogonalization into these algorithms,
all of which increase algorithmic costs.
This motivates reorthogonalizing selectively.
We briefly investigated using diagnostics based on $U^TU$ and $V^TV$,
monitored periodically to amortize the cost of these matrix multiplications,
and believe this is a practical way forward.

\subsection{Decoherence of Momentum}
\label{sec:momentum-decoherence}

Our algorithms represent momentum using CD-coordinates in the fixed-rank embedded algorithm and the singleton partial-isometry embedded algorithm,
and using a horizontal lift in the other algorithms.
We call such representation coherent if it actually represents a tangent vector at the current base point
and, in the horizontal lift case, in the intended gauge.
Coherence is automatic in exact arithmetic but can be lost in finite arithmetic,
potentially invalidating our norm computations, retractions, and vector transports.
We propose repairing the momentum immediately after updating it with the new gradient,
before consuming it to compute the norm, etc.

In the cases that use CD-coordinates
--- the fixed-rank embedded and singleton partial isometry embedded algorithms ---
coherence means that $U^T C = D^T V$,
and,
in the partial isometry case,
additionally that this matrix is skew-symmetric.
Here,
$U,V$ are components of the USV-representation of $W$ used in the current step of the algorithm,
(i.e., a UV-representation in the partial isometry case),
and $C,D$ are the provisional CD-coordinates of the momentum $M$.
We propose seeking a tangent vector that solves the least-squares problem
\[
\min_\Xi \|\Xi V - C\|_F^2 + \|\Xi^T U - D\|_F^2
\text.
\]
In either case,
there is a unique minimum-norm solution $M^*$,
whose CD-coordinates are
\begin{align*}
C^* &= C - U(K_C - K^*)
\\
D^* &= D - V(K_D - K^*)^T
\text,
\end{align*}
where $K_C = U^TC$, $K_D = D^T V$, and $K^* = (K_C + K_D)/2$ in the fixed-rank case,
or $K^* = \Skew(K_C + K_D)/2$ in the partial isometry case.
The current implementation takes a cheaper approach:
it uses the aforementioned averages $K^*$ when converting to XKY-coordinates,
but doesn't repair the momentum itself,
so, for example, $X$ and $Y$ may not end up orthogonal to $U$ and $V$.

In the other algorithms,
which represent momentum as a horizontal vector,
the concern is loss of tangency and horizontality,
which motivates us to seek the horizontal vector that solves the least-squares problem
\[
\min_{(\Xi_A,\Xi_B)} \|(\Xi_A,\Xi_B) - (M_A,M_B)\|^2_{\bar g}
\text.
\]
In all cases,
a unique minimum-norm solution $(M^*_A, M^*_B)$ exists,
and is obtained by tangent and horizontal projection.
For the fixed-rank quotient geometry (handling the grid case by stacking),
\[
(M^*_A,M^*_B)
=
(M_A-A\Lambda,\, M_B+B\Lambda^T)
\text,
\]
where
\[
\Lambda
= \frac12\left((A^TA)^{-1}A^TM_A - M_B^T B(B^TB)^{-1}\right)
\text.
\]

For the grid partial-isometry quotient and canonical algorithms,
and for the grid partial-isometry embedded algorithm once it has gone upstairs,
the repair is factorwise projection onto the Stiefel tangent spaces then removal of the vertical component:
\begin{gather*}
\begin{aligned}
M'_{U_i} &= M_{U_i} - U_i \Sym(M_{U_i}^T U_i)
\text,
&
M'_{V_j} &= M_{V_j} - V_j \Sym(M_{V_j}^T V_j)
\text,
\end{aligned}
\\
\Omega = \frac{J\sum_i U_i^T M'_{U_i} + I\sum_j V_j^T M'_{V_j}}{2IJ}
\text,
\\
\begin{aligned}
M^*_{U_i} &= M'_{U_i} - U_i\Omega
\text,
&
M^*_{V_j} &= M'_{V_j} - V_j\Omega
\text.
\end{aligned}
\end{gather*}
For the singleton cases, take $I=J=1$.

Of course, the cheapest possible mitigation is a hard reset, $M\gets 0$.
This may be the most appropriate approach for large defects.

These repairs can be applied every iteration,
periodically, or
only when such diagnostics exceed a tolerance.
For CD-coordinates, we suggest tracking
decoherence, $\|U^TC-D^TV\|_F$,
and, in the partial isometry case, also the deskewing,
$\|\Sym(U^TC)\|_F + \|\Sym(D^TV)\|_F$.
For the other methods, we suggest monitoring norms of the tangent and horizontal residuals.
We suspect that decoherence is much less of an issue for these other methods,
because our implementations already perform these repairs as part of vector transport;
our biggest concern there is decoherence introduced by the provisional lifted gradient.

\subsection{Gradient Inconsistency}
\label{sec:impl-recover-gradient}

Let us expand on our statements in \cref{sec:recover-GW} and \cref{sec:separable-loss}.

An ambient Euclidean gradient $G_W$ exists if and only if $G_A = G_A B^+B$, $G_B = G_B A^+ A$, and $A^T G_A = G_B^TB$,
in which case
\[
G_W \in \{ G_W^* + (I-AA^+) K (I-BB^+) : K \in \reals^{m \times n}\}
\text,
\]
where the minimum Frobenius-norm version
\begin{equation}
\label{eq:gstar}
G^*_W
= A^{+T} G_B^T + G_A B^+ - A A^+ G_A B^+
= A^{+T} G_B^T + G_A B^+ - A^{+T} G_B^T B B^+
\text.
\end{equation}
If the objective $f$ depends only on $W=AB^T$,
not the factors $A$ and $B$ separately,
then the three consistency conditions are automatically satisfied (in exact arithmetic).
We made this assumption throughout, and discussed some implications in \cref{sec:separable-loss}.
Moreover,
we assumed both $A$ and $B$ are full rank,
thus $A^T G_A = G_B^TB$ is the only nontrivial condition;
we discussed some obstacles to relaxing this in \cref{sec:bounded-rank}.
Lastly,
non-uniqueness turns out to be innocuous:
in all cases,
we will obtain the same Riemannian gradient $G^R_W$ regardless of which version of $G_W$ we use.

As mentioned in \cref{sec:separable-loss},
the preceding discussion also applies to the partial isometry case,
substituting $U,V$ for $A,B$,
and to the grid case (\cref{sec:algs-grid}),
substituting $\cA,\cB,\cW$ for $A,B,W$.
And in the partial isometry case,
we technically don't actually need $G_W$ to exist in order to recover the Riemannian gradient:
for this we only need the weaker condition,
that $\Skew(U^T G_U) = \Skew(G_V^T V)$ (singleton)
or $\Skew(\cU^T G_\cU) = \Skew(G_\cV^T \cV)$ (grid).

When the consistency condition fails,
the Riemannian gradient is not well defined,
and the quantities produced within our algorithms should not be trusted.
We discuss a few mitigations.

In the fixed-rank embedded case,
we seek a tangent vector $\Xi$ that solves the least-squares problem,
\[ \min_\Xi \| \Xi B - G_A\|_F^2 + \| \Xi^T A - G_B \|_F^2\text. \]
There exists a unique minimum-norm solution $\Xi^*$,
whose CD-coordinates are given by
\begin{align*}
C^* &= C - U(K_C - K^*)
\\
D^* &= D - V(K_D - K^*)^T
\text,
\end{align*}
where
$C = G_A R_B^{-1}$,
$D = G_B R_A^{-1}$,
$A = U R_A$ and $B=V R_B$ are thin QR factorizations,
$K_C = U^T C$,
$K_D = D^T V$, and
$K^*$ is the unique solution to the Sylvester equation
\[
S_A K^* + K^* S_B = S_A K_D + K_C S_B
\text,
\]
which can be found in $O(r^3)$ operations by the standard method
involving spectral decompositions of the symmetric positive definite matrices $S_A = R_A R_A^T$ and $S_B = R_B R_B^T$.
In the grid case, just replace $A,B,W$ by $\cA,\cB,\cW$ in the preceding.

A cheaper repair,
in the fixed-rank embedded case,
is to average the overlap.
This leads to the same update formula as above,
except with
$K^* = (K_C + K_D)/2$.
This applies to the grid case as well.
At the time of writing,
this cheaper safeguard was commented out in our implementation,
so the only protection against gradient inconsistency was the aforementioned averaging performed during XKY-coordinate conversion.

For the singleton partial isometry embedded case,
the preceding least-squares problem can be solved much more cheaply.
The solution is the same as above,
substituting $A,B$ by $U,V$ and $R_A,R_B,S_A,S_B$ by $I$,
and computing $K^* = \Skew(K_C + K_D)/2$.
That is,
the Sylvester solve becomes trivial (like in the case of momentum repair).
The averaging technique in the current implementation amounts to an incomplete version of this repair,
that does not guarantee the resulting CD-coordinates are coherent.
(A similar partial mitigation is performed shortly thereafter, in the XKY-coordinate conversion.)

To nobody's surprise,
the grid partial isometry embedded case is once again much more complicated.
We will describe an approximate analogue of the preceding principled least-squares approach,
although we suspect it may be too expensive to be useful in practice.
The basic idea is to first obtain the repaired gradient $\Xi^*$ for the grid fixed-rank embedded case,
following the discussion above,
then replace its grid K-coordinates by the nearest ones satisfying the additional skew-symmetry requirements of the grid partial isometry case.
Continuing notation from above,
but omitting the stars from $\Xi^*$ and $K^*$ to avoid confusion
--- we haven't finished repair yet ---
we have that
\begin{align*}
\Xi
&= (I-UU^T)C V^T
+U D^T(I-VV^T)
+UKV^T
\\
&=
C V^T+ U D^T + U(K-K_C-K_D)V^T
\text;
\end{align*}
this form allows extracting blocks from $\Xi$ without materializing the whole matrix.
For each block $i,j$,
define
\[
H_{ij} = \Skew(U_i^T \Xi_{ij} V_j)
\text.
\]
Now,
recall that,
in grid-XKY coordinates in the partial isometry case,
the K-factors aren't arbitrary skew-symmetric matrices,
so we can't just use $H_{ij}$ as $K^*_{ij}$.
Rather,
there must exist skew-symmetric matrices $\Omega_i$ and $\Psi_j$ such that
$K^*_{ij}=\Omega_i-\Psi_j$.
This motivates us to solve the least squares problem,
\[
\min_{(\Omega_i)_i,(\Psi_j)_j}
\sum_{i,j} \| H_{ij} - (\Omega_i - \Psi_j)\|_F^2
\text.
\]
For any optimum $((\Omega^*_i)_i, (\Psi^*_j)_j)$,
for all $i,j$,
\[
K^*_{ij} = \Omega^*_i - \Psi^*_j = \frac1J \sum_\ell H_{i\ell} + \frac1I \sum_k H_{kj} - \frac1{IJ}\sum_{k,\ell}H_{k\ell}
\text,
\]
and the matrices $(K^*_{ij})_{ij}$ are the same for every optimum.
We take
\begin{align*}
\Omega_i^* &= \frac1J \sum_\ell H_{i\ell} - \frac1{2IJ}\sum_{k,\ell}H_{k\ell}
\text,
\\
\Psi_j^* &= \frac1{2IJ}\sum_{k,\ell}H_{k\ell} - \frac1I \sum_k H_{kj}
\text,
\end{align*}
because they satisfy the horizontal condition
$J\sum_i \Omega^*_i + I\sum_j \Psi^*_j = 0$.
The repaired gradient is then obtained via its grid XKY-coordinates,
\[
\Xi^*_{ij} = X_i^* V_j^T + U_i {Y_j^*}^T + U_i K^*_{ij} V_j^T
\text,
\]
with these $K^*_{ij}$ and
\begin{align*}
X_i^* &= \frac1J \sum_j (I-U_iU_i^T) \Xi_{ij} V_j
\text,
&
Y_j^* &= \frac1I \sum_i (I-V_jV_j^T) \Xi_{ij}^T U_i
\text,
\end{align*}
Finally, the horizontal lifts are
\[
G^{R,*}_{U_i}=X_i^* + U_i\Omega_i^*
\text,
\qquad
G^{R,*}_{V_j}=Y_j^* + V_j\Psi_j^*
\text.
\]
Concerned that this may be too expensive,
we implemented a cheaper approximation,
where we apply the same projection only to the row and column skew summaries.
That is,
letting
\[
\Alpha_i = \Skew(U_i^T G_{U_i})
\text,
\qquad
\Beta_j = \Skew(G_{V_j}^T V_j)
\text,
\qquad
\Gamma = \frac12\left(\sum_i \Alpha_i + \sum_j \Beta_j\right)
\text,
\]
we take
\[
\Omega_i = \frac1J\Alpha_i - \frac1{2IJ}\Gamma
\text,
\qquad
\Psi_j = \frac1{2IJ}\Gamma - \frac1I\Beta_j
\text,
\]
which is exact when gradients are consistent,
and otherwise gives a low-cost projection of their inconsistent skew summaries.

In the fixed-rank quotient geometry,
the principled repair is to horizontally project the provisional (automatically tangent) quotient gradient.
Similarly,
for the partial-isometry quotient and canonical geometries,
the analogous repair is to tangent-project the provisional lifted factor-space direction
and then apply the appropriate horizontal projection.
In all of these cases, the result is the closest horizontal direction in the upstairs metric.

In the preceding,
we assumed that the gradient inconsistency is minor, arising, e.g., from floating-point rounding errors.
If the objective genuinely depends on the factors separately,
these approaches should instead be interpreted as yielding a heuristic search direction.

\printbibliography[title={References for Appendices}]
\end{refsection}

\end{document}